\documentclass[10pt,journal,compsoc]{IEEEtran}
\ifCLASSOPTIONcompsoc
  \usepackage[nocompress]{cite}
\else
  \usepackage{cite}
\fi

\ifCLASSINFOpdf

\else

\fi

\usepackage{times}
\usepackage{epsfig}
\usepackage{graphicx}
\usepackage{amsmath}
\usepackage{amssymb}
\usepackage{booktabs}
\usepackage{multirow}
\usepackage{xcolor}
\usepackage{subfigure}
\usepackage{colortbl}
\definecolor{tabcolor}{rgb}{ 0.5,  0.5,  0.5}
\usepackage[colorlinks,linkcolor=blue]{hyperref}

\newcommand{\etal}{et al. }
\newcommand{\ie}{i.e., }
\newcommand{\eg}{e.g., }
\newcommand{\etc}{etc}
\usepackage{threeparttable}
\usepackage{footnote}

\begin{document}
%
\title{Background-Click Supervision for \\ Temporal Action Localization}

\author{Le~Yang, Junwei~Han,~\IEEEmembership{Senior~Member,~IEEE,} Tao~Zhao, Tianwei~Lin \\ Dingwen~Zhang~\IEEEmembership{Member,~IEEE} and Jianxin Chen
\IEEEcompsocitemizethanks{\IEEEcompsocthanksitem This work was supported in part by the Key-Area Research and Development Program of Guangdong Province (2019B010110001), in part by the National Natural Science Foundation of China under Grants 61876140, in part by the Innovation Foundation for Doctor Dissertation of Northwestern Polytechnical University under Grant CX201916. (\emph{Corresponding authors: Junwei Han and Dingwen Zhang.})
\IEEEcompsocthanksitem L. Yang, J. Han, T. Zhao, and D. Zhang are with the Brain Lab (\textcolor[rgb]{0.9255,0.0078,0.5529}{https://nwpu-brainlab.gitee.io/index\_en.html}), Northwestern Polytechnical University. T. Lin is with Baidu VIS. Jianxin Chen is with Beijing University of Chinese Medicine. (e-mails: junweihan2010@gmail.com and zhangdingwen2006yyy@gmail.com).}
}

\markboth{IEEE TRANSACTIONS ON PATTERN ANALYSIS AND MACHINE INTELLIGENCE, VOL. -, NO. -, - 2021}%
{Shell \MakeLowercase{\textit{et al.}}: Bare Demo of IEEEtran.cls for Computer Society Journals}

\IEEEtitleabstractindextext{%
\begin{abstract}
		Weakly supervised temporal action localization aims at learning the instance-level action pattern from the video-level labels, where a significant challenge is action-context confusion. To overcome this challenge, one recent work builds an action-click supervision framework. It requires similar annotation costs but can steadily improve the localization performance when compared to the conventional weakly supervised methods. In this paper, by revealing that the performance bottleneck of the existing approaches mainly comes from the background errors, we find that a stronger action localizer can be trained with labels on the background video frames rather than those on the action frames. To this end, we convert the action-click supervision to the background-click supervision and develop a novel method, called BackTAL. Specifically, BackTAL implements two-fold modeling on the background video frames, i.e. the position modeling and the feature modeling. In position modeling, we not only conduct supervised learning on the annotated video frames but also design a score separation module to enlarge the score differences between the potential action frames and backgrounds. In feature modeling, we propose an affinity module to measure frame-specific similarities among neighboring frames and dynamically attend to informative neighbors when calculating temporal convolution. Extensive experiments on three benchmarks are conducted, which demonstrate the high performance of the established BackTAL and the rationality of the proposed background-click supervision. Code is available at \href{https://github.com/VividLe/BackTAL}{https://github.com/VividLe/BackTAL}.
\end{abstract}

}

\maketitle

\IEEEdisplaynontitleabstractindextext

\IEEEpeerreviewmaketitle

\IEEEraisesectionheading{\section{Introduction}\label{sec:introduction}}

\IEEEPARstart{T}{emporal} action localization aims at discovering action instances via predicting the corresponding start times, end times, and action category labels \cite{gaidon2013temporal}. It is a challenging yet practical research topic, with potential benefits to a wide range of intelligent video processing systems, \eg video summary \cite{zhao2019property}, smart surveillance \cite{ko2018deep}. To achieve precise localization performance, the fully supervised temporal action localization methods \cite{xu2017r, chao2018rethinking, zeng2019graph,xu2019two} learn from human annotations. However, the data annotation process is burdensome and expensive; meanwhile, it is difficult to consistently determine the boundary of action among different annotators. In contrast, the weakly supervised methods \cite{wang2017untrimmednets,nguyen2018weakly, nguyen2019weakly} learn from the video-level category labels, which are cheap and convenient to obtain.

\begin{figure}[t]
	\graphicspath{{figure/}}
	\centering
	\includegraphics[width=1\linewidth]{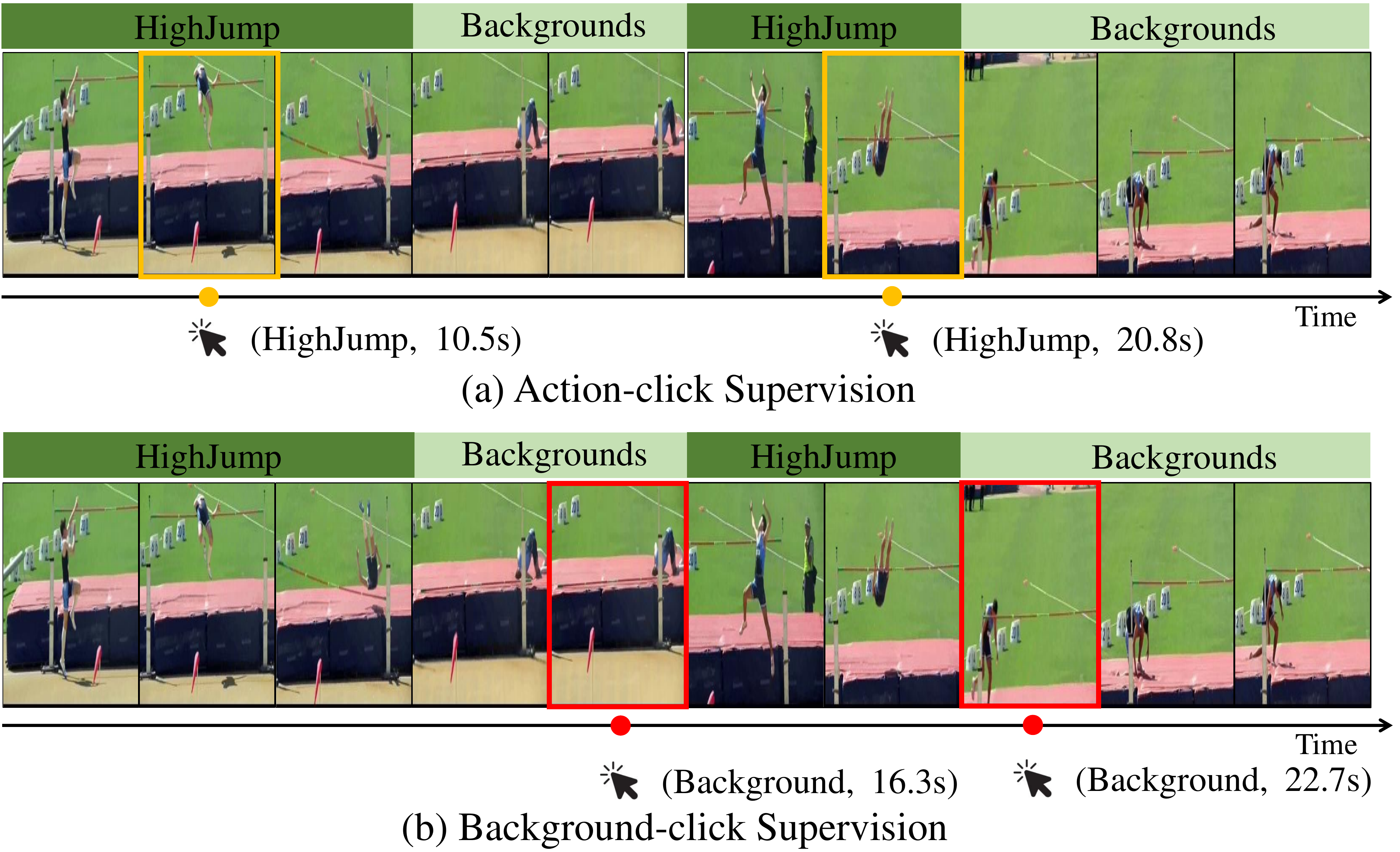}
	\caption{Illustration of click-level supervision. The ground truth is shown above the frame sequence. (a) The action-click supervision (shown in orange) makes a random click within each action instance, records the timestamp and classification label, used by SF-Net \cite{ma2020sf}. (b) The proposed background-click supervision (shown in red) makes a random click within each background segment and record the timestamp.}
	\label{motivation}
\end{figure}

Inherently, most weakly supervised algorithms follow an underlying assumption that video segments contributing more evidence to video-level classification are more likely to be action. Nonetheless, the algorithm developed following this assumption would be stuck into an action-context confusion dilemma when background segments are more related to video-level classification, as pointed out by Shi \etal \cite{shi2020weakly} and Choe \etal \cite{choe2020evaluating}. Recently, SF-Net \cite{ma2020sf} enhances the weakly supervised algorithms via introducing the action-click supervision\footnote{The action-click supervision is termed as single-frame supervision in SF-Net \cite{ma2020sf}. To indicate one timestamp is clicked within each action instance, we term this supervision as ``action-click supervision".}, as shown in Fig. \ref{motivation}(a). They randomly click a timestamp within each action instance and annotate the corresponding category label. In this work, Ma \etal \cite{ma2020sf} show that given a one-minute video, making video-level, click-level, and instance-level annotation require 45s, 50s, and 300s, respectively. Specifically, it requires watching the whole video when annotating the video-level category label. Similarly, annotating the action-click supervision requires watching the whole video and randomly click an action frame once encountering an action segment. Because the click information is automatically generated by the annotation tool, the action-click supervision costs similar annotation time with the video-level supervision. However, the instance-level annotation needs to roll back and forth to precisely determine the starting frame and the ending frame, whose annotation cost dramatically increases. Because the cost to annotate the click-level supervision is affordable, meanwhile, as verified in experiments, click-level supervision can indicate the action frame and partially mitigate the action-context confusion challenge, weakly supervised temporal action localization with the click-level annotation exhibits promising research prospect.

\begin{figure}[t]
	\graphicspath{{figure/}}
	\centering
	\includegraphics[width=1\linewidth]{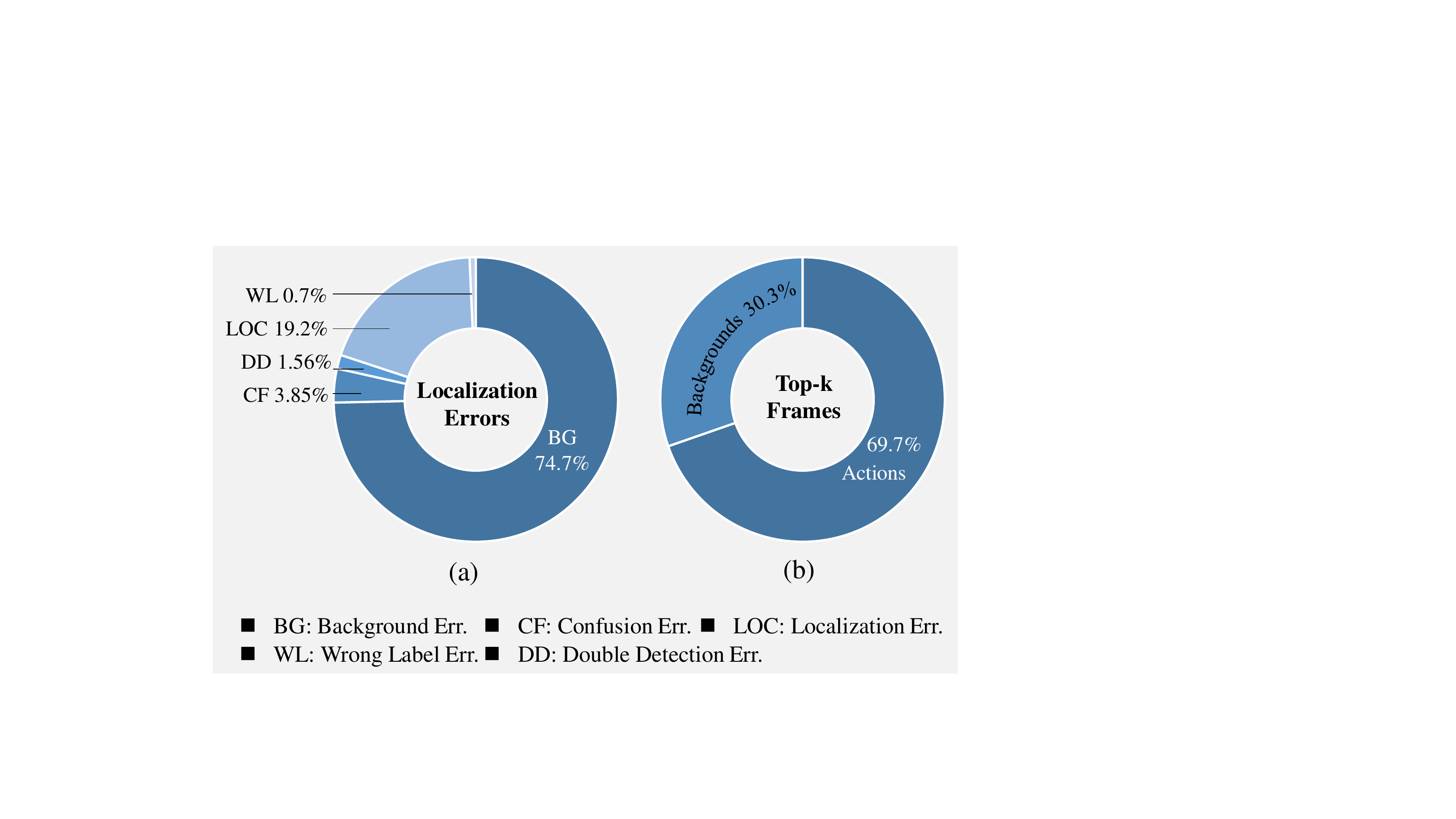}
	\caption{Performance analysis of a baseline method for weakly supervised action localization. (a) From the diagnosing results, it can be found that the majority of errors come from Background Error. (b) Given class activation sequence, the majority of top-$k$ frames fall into action segments.}
	\label{bg-superiority}
\end{figure}

Although SF-Net \cite{ma2020sf} has advanced the localization performance by exploring action-click supervision, it is still questionable whether the click-level supervision on background segments would perform better. Specifically, besides the video-level classification label, we can click a frame within each background segment (see Fig. \ref{motivation}(b)). To study this, we implement a baseline method by following previous weakly supervised methods \cite{lee2020background, liu2019completeness, min2020adversarial, zhai2020two} and carry two experiments on THUMOS14. Specifically, we forward video features through three temporal convolutional layers and predict the classification score for each frame. The experimental results are reported in Fig. \ref{bg-superiority}. We employ the diagnosing tool \cite{alwassel2018diagnosing} to perform error analysis. Among five types of errors, the vast majority of errors come from the Background Error and take up $74.7\%$, as shown in Fig. \ref{bg-superiority}(a). On the contrary, a part of action frames can be confidently determined via the top-$k$ aggregation procedure. Specifically, given each frame's classification score, the existing paradigm usually selects the highest $k$ scores and regards the mean value as the video-level classification score. Consequently, a well-trained model can confidently discover reliable action frames according to selected top-$k$ frames. We measure the ratio that the highest $k$ scores fall into action segments and obtain $69.7\%$, as shown in Fig. \ref{bg-superiority}(b). The diagnosing results and the distribution of top-$k$ frames inspire us to convert the action-click supervision into the background-click supervision.

\begin{figure*}[t]
	\graphicspath{{figure/}}
	\centering
	\includegraphics[width=1\linewidth]{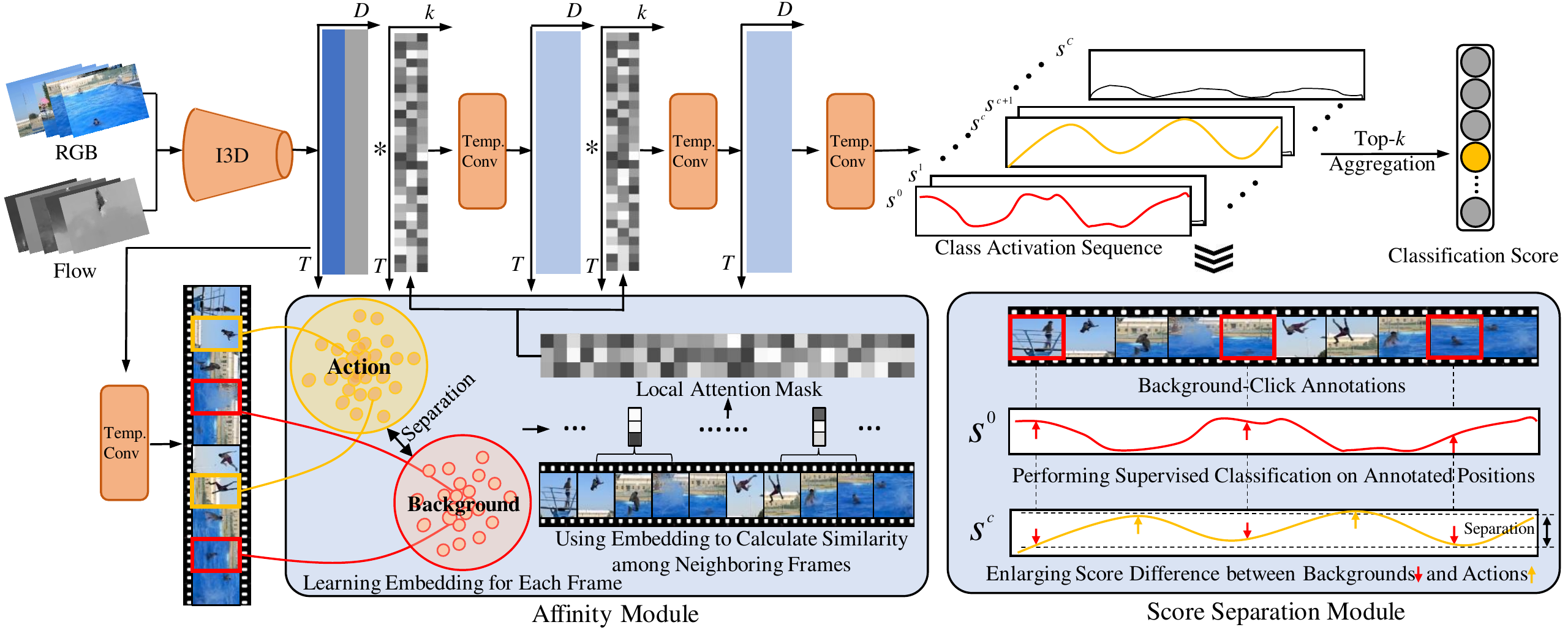}
	\caption{Framework of the proposed BackTAL. BackTAL first extracts video features, then uses three temporal convolutional layers to classify each frame and obtain the class activation sequence. Finally, it performs top-$k$ aggregation and predicts the video-level classification score. Based on the background-click supervision, BackTAL adopts the affinity module (see section \ref{affinity-module}) to mine the feature information and estimate the local attention mask. This assists in calculating frame-specific temporal convolution. Besides, BackTAL explores the score separation module (see section \ref{score-separation}) and mines the position information. This can enlarge the response gap between action frames and background frames.}
	\label{fig-framework}
\end{figure*}

To effectively learn the action localization model, we propose a novel learning framework under background-click supervision, called BackTAL, as shown in Fig. \ref{fig-framework}. Given background-click supervisions, a direct way to leverage the annotation information is to mine the position information via performing the supervised classification on the annotated frames, which is principally explored by SF-Net \cite{ma2020sf} by performing category-specific classification and category-agnostic classification. Besides, considering the commonly used top-$k$ aggregation procedure only constrains the highest $k$ scores but ignores other positions, we design a \textit{Score Separation Module}. It strives to enlarge the score differences between potential action frames and annotated background frames, which can thoroughly mine the position information and further improve the localization performance.

In addition to position information, click-level supervision can also guide the process to build feature embedding spaces that separate action features from background features. However, this feature information is ignored by previous work \cite{ma2020sf}. We propose an \textit{Affinity Module} to utilize the feature information and realize the dynamic and precise calculation for the temporal convolutional operation. Given annotated background frames and confident action frames, the affinity module strives to learn an appropriate embedding for each frame by requiring all action embeddings to be compact, all background embeddings to be compact, while action embeddings to preserve considerable distance from background embeddings. Whereafter, high-quality embeddings are used to estimate similarities between a frame and its local neighboring frames, generating similarity masks. Assisted with the proposed frame-specific similarity masks, the convolution kernel can dynamically attend to relevant neighboring frames when performing calculation on each frame, achieving precise calculation for the temporal convolution.

Our contributions can be summarized as follows:

\begin{itemize}
	\item We propose background-click supervision for the temporal action localization task. Compared with the existing action-click supervision, it can effectively discover action instances but with similar annotation costs.
	\item In BackTAL, we propose a score separation module and an affinity module to endow the simple yet effective modeling of the position information and the feature information, respectively.
	\item Extensive experiments are performed on three benchmarks, and BackTAL achieves new high performances, \eg 36.3 with mAP@tIoU0.5 on THUMOS14.
\end{itemize}

The rest of this paper is organized as follows. Section \ref{sec-related-work} reviews recent progresses in temporal action localization under both full supervision and weak supervision, as well as the development of the click-level supervision. Then, section \ref{sec-method} presents the proposed BackTAL in details, including the holistic method, the score separation module and the affinity module. Afterwards, experimental results are exhibited in section \ref{sec-experiments}. Specifically, we introduce the background-click annotation process, perform comparison experiments on three benchmark datasets, and carry ablation studies to analyze the effectiveness of each module, in quantitative and qualitative manner. Finally, section \ref{sec-conclusion} draws the conclusion and discusses further potential works.

\section{Related Work}
\label{sec-related-work}
This section summarizes recent progresses about the temporal action localization task \cite{gaidon2013temporal, lee2021weakly, zhang2021weakly}. We start from fully supervised methods, and review one-stage methods and two-stage methods. Then, we discuss weakly supervised methods that only learn from video-level classification labels. In the end, we discuss an enhanced weakly supervised learning paradigm, \ie click-level supervision.

\textbf{Fully supervised temporal action localization} methods learn from precise annotation for each frame. Existing works can be summarized into one-stage methods \cite{lin2017single, long2019gaussian, xu2020g} and two-stage methods \cite{xu2017r, chao2018rethinking, lin2019bmn, zeng2019graph,xu2019two}. For the former type, Lin \etal \cite{lin2017single} simultaneously predict action boundaries and labels, which is developed by GTAN \cite{long2019gaussian} via exploiting gaussian kernels. Recently, Xu \etal \cite{xu2020g} employ a graph convolutional network to perform one-stage action localization. In contrast, two-stage methods first generate action proposals, then refine and classify confident proposals. Specifically, a majority of proposals are generated by the anchor mechanism \cite{xu2017r, gao2017turn, chao2018rethinking,xu2019two, yang2020revisiting}. In addition, other ways to generate proposals include sliding window \cite{shou2016temporal}, temporal actionness grouping \cite{zhao2017temporal}, combining confident starting and ending frames \cite{lin2018bsn, lin2019bmn}. Afterwards, MGG\cite{liu2019multi} integrates the anchor mechanism and frame actionness mechanism into a unified framework, which achieves high recall and precision for the proposal generation task. Unlike fully supervised methods, the studied weakly supervised setting lacks precise instance-level annotations, leaving distinguishing actions from backgrounds challenging.

\textbf{Weakly supervised temporal action localization} chiefly mines video-level classification labels. The pioneering works UntrimmedNet \cite{wang2017untrimmednets}, STPN \cite{nguyen2018weakly}, and AutoLoc \cite{shou2018autoloc} build the paradigm that localizes action instances via thresholding the class activation sequence. Whereafter, the video-level category label is thoroughly mined, \eg W-TALC \cite{paul2018w} explores the co-activity similarity between two videos sharing the same label, which inspires Gong \etal \cite{gong2020learning} to mine co-attention features. Besides, both Liu \etal \cite{liu2019completeness} and Min \etal \cite{min2020adversarial} demonstrate that learning multiple parallel and complementary branches is beneficial to generate complete action instances, which is developed by HAM-Net \cite{islam2021a} that learns hybrid attention weights to localize complete action instances. Moreover, CleanNet \cite{liu2019weakly} proposes a pseudo supervision paradigm, which firstly generates pseudo action proposals, and then employs these proposals to train an action proposal network. The pseudo supervision paradigm is further developed by some recent works, such as TSCN \cite{zhai2020two} and EM-MIL \cite{luo2020weakly}. In addition, BaS-Net \cite{lee2020background} designs a background suppression network, which is developed by Moniruzzaman \etal \cite{moniruzzaman2020action} via further modeling action completeness. Similarly, Nguyen \etal \cite{nguyen2019weakly} point out that it is crucial to model backgrounds. This is summarized as the action-context confusion challenge by \cite{shi2020weakly} and \cite{zhao2021soda}. Recently, Lee \etal \cite{lee2021weakly} study frame's inconsistency and model background frames as out-of-distribution samples. Meanwhile, Liu \etal \cite{liu2021acsnet, liu2021weakly} aim to separate action frames and neighboring context frames via employing the positive component and negative component \cite{liu2021acsnet}, or learning explicit subspace \cite{liu2021weakly}. 

However, the action-context confusion challenge is far from solved if only using video-level labels. In contrast, introducing extra information may be a more effective solution. For example, CMCS \cite{liu2019completeness} collects stationary video clips as backgrounds. Besides, 3C-Net \cite{narayan20193c} introduces action count cues. Nguyen \etal \cite{nguyen2019weakly} employs micro-videos. Recently, ActionBytes \cite{jain2020actionbytes} learns from trimmed videos and localizes actions in untrimmed videos.

\textbf{Click-level supervision} is a kind of weakly supervised learning paradigm. As Bilen \cite{Bilen14wsol} points out, weakly supervised learning refers to an algorithm that requires cheaper annotations at the training phase than the desired output at the inference phase. For example, from point supervision to pixel-level semantic mask \cite{bearman2016s}, from points at frames to spatio-temporal action boxes \cite{mettes2016spot}, from scribble to pixel-level segmentation mask \cite{lin2016scribblesup} and saliency maps \cite{zhang2020weakly}. Recently, Moltisanti \etal \cite{moltisanti2019action} employ simulated action-click supervision to learn video recognition models.

Compared with the most relevant work, SF-Net \cite{ma2020sf}, our proposed BackTAL exhibits two distinguishable contributions. (1) Although SF-Net uses the action-click annotation, we find that action frames can be confidently discovered by the learning algorithm, while the performance bottleneck lies in background errors. Thus, we convert the action-click annotation to the background-click annotation. (2) Given the click-level annotation, SF-Net principally mines the position information via supervised classification on the annotated frame, while we jointly explore the position information and the feature information via the score separation module and the affinity module.

The proposed affinity module is related to embedding learning \cite{harley2017segmentation, liu2020picanet, ci2018video}. In detail, PiCANet \cite{liu2020picanet} directly learns affinity among neighboring pixels, while BackTAL learns embedding for each frame and then measures affinity. Moreover, existing methods \cite{harley2017segmentation, liu2020picanet, ci2018video} learn embedding under a fully supervised setting, while BackTAL learns from the background-click annotation.

In addition to temporal action localization works discussed above, there is a similar task temporal action segmentation, which receives promising advances recently. For example, MS-TCN++ \cite{li2020ms} utilizes a multi-stage architecture to tackle the temporal action segment task. Kuehne \etal \cite{kuehne2018hybrid} learn action classifiers from videos with action order information, and integrate a framewise RNN model with hidden Markov model to segment video. In addition, some works \cite{tran2013video,soomro2018online, su2020progressive} study the spatio-temporal action detection, which detects action instances via spatial boxes within each temporal frame.

\section{Method}
\label{sec-method}
In this section, we elaborate the proposed BackTAL method to tackle the weakly supervised temporal action localization under background-click supervision. First, section \ref{sec-problem-statement} formally defines the studied problem. Then, a holistic overview is presented in section \ref{sec-overview}, where we introduce the traditional video-level classification loss and the frame-level classification loss to mine the position information. Given background-click supervision, BackTAL simultaneously mines the position information and the feature information. The former is depicted in section \ref{score-separation} while the latter is depicted in section \ref{affinity-module}. Afterwards, the evaluation process is introduced in section \ref{method-action-localization}.

\subsection{Problem Statement}
\label{sec-problem-statement}
The proposed BackTAL tackles untrimmed videos via learning from video-level classification labels and background-click annotations. Given a video, we denote the background-click label as $\mathbf{b}=[b_{1}, b_{2},...,b_{T}]$. Before the human annotation process, the background-click labels for all frames are $b_{t}=-1, t=1,...,T$, indicating that it is uncertain whether each frame belongs to action or background. In the annotation process, the annotator goes through all video frames. Once the annotator encounters a series of consecutive background frames, he/she randomly selects a frame and makes the background-click annotation, \ie marking the corresponding background-click label as $b_{t}=1$.\footnote{Please refer to Subsection \textit{4.2 Background-Click Annotation} for a detailed annotation process.} During training, the algorithm selects the highest $k$ scores to estimate the video-level classification score, which is called the top-$k$ aggregation procedure. We regard the selected $k$ frames as confident action frames, and mark the corresponding label as $b_{t}=0$, then obtain pseudo label $\hat{\mathbf{b}}$. BackTAL learns from training videos and aims to precisely discover action instances, \eg  $(t^{s}_{i},t^{e}_{i},c_{i},p_{i})$, in the testing videos. Specifically, the $i^{th}$ action instance starts at $t^{s}_{i}$, ends at $t^{e}_{i}$, belongs to the $c^{th}$ class, and the confidence score for this prediction is $p_{i}$.

\subsection{BackTAL Overview}
\label{sec-overview}
The framework of BackTAL is shown in Fig. \ref{fig-framework}. BackTAL employs three temporal convolutional layers to dispose video feature sequences, to perform classification for each frame and to generate the class activation sequence. For the input video with feature $\mathbf{X}$, the corresponding class activation sequence is $\mathbf{S} \in \mathbb{R}^{(C+1) \times T}$. Afterwards, we  utilize the top-$k$ aggregation strategy to calculate the video-level classification score $s^{c}_{v}$:
\begin{equation}
s^{c}_{v}=\frac{1}{k} \max _{\mathcal{M} \subset \mathbf{S}[c,:] \atop |\mathcal{M}|=k} \sum_{l=1}^{k} \mathcal{M}_{l},
\end{equation}
where $s^{c}_{v}$ is the classification score for the $c^{th}$ class.\footnote{The temporal locations of the selected highest $k$ scores can be denoted as a set $\mathcal{K}=\{k_{1}, k_{2}, ..., k_{k}\}$, ($k_{i}\in\{1,2,...,T\}$), where the corresponding pseudo frame-level label satisfies $\hat{b}_{t}=0, t \in \mathcal{K}$.}

Given video-level classification score $\mathbf{s}_{v}=[s^{0}_{v},s^{1}_{v},...,s^{C}_{v}]$ and classification label $\mathbf{y}$, the video-level classification loss $\mathcal{L}_{\rm cls}$ can be calculated via the cross-entropy loss:
\begin{equation}
	\mathcal{L}_{\rm cls} = - \sum_{c=0}^{C} y^{c} {\rm log}(\hat{s}^{c}_{v}),
\end{equation}
where $\mathbf{\hat{s}}_{v}=[\hat{s}^{0}_{v},\hat{s}^{1}_{v},...,\hat{s}^{C}_{v}]$ is the classification score after softmax normalization.

In addition to video-level classification, we perform supervised classification on annotated background frames to improve the quality of the class activation sequence $\mathbf{S}$. Specifically, consider an annotated frame with label $b_{t}=1$, this frame's classification score is $\mathbf{S}[:,t] \in \mathbb{R}^{(C+1) \times 1}$. We first perform softmax normalization and obtain the frame-level classification score $\mathbf{\hat{s}}_{t}=[\hat{s}^{0}_{t},\hat{s}^{1}_{t},...,\hat{s}^{C}_{t}]$. Then, we calculate the frame-level classification loss $\mathcal{L}_{\rm frame}$ via the cross-entropy loss:
\begin{equation}
	\mathcal{L}_{\rm frame} = - \frac{1}{N_{\rm frame}} \sum_{t=1}^{N_{\rm frame}} {\rm log}(\hat{s}^{0}_{t}),
\end{equation}
where $N_{\rm frame}$ is the number of annotated background frames within this video, and $\hat{s}^{0}_{t}$ is the classification score for the background class.

During training, background frames are annotated, and the highest $k$ scores of the class activation sequence can be regarded as confident action frames. In the Score Separation Module (see section \ref{score-separation}), we aim to separate response scores between confident action frames and annotated background frames via the separation loss $\mathcal{L}_{\rm sep}$. In the Affinity Module (see section \ref{affinity-module}), we learn embedding for each frame via the affinity loss $\mathcal{L}_{\rm aff}$, and employ embedding vectors to measure similarities among neighboring frames.

The complete learning process is jointly driven by video-level classification loss $\mathcal{L}_{\rm cls}$, frame-level classification loss $\mathcal{L}_{\rm frame}$, separation loss $\mathcal{L}_{\rm sep}$ and affinity loss $\mathcal{L}_{\rm aff}$. The total loss can be calculated as:
\begin{equation}
\mathcal{L}=\mathcal{L}_{\rm cls}+\mathcal{L}_{\rm frame}+\lambda \cdot \mathcal{L}_{\rm sep} + \beta \cdot \mathcal{L}_{\rm aff},
\end{equation}
where $\lambda$ and $\beta$ are trade-off coefficients.

\begin{figure}[t]
	\graphicspath{{figure/}}
	\centering
	\includegraphics[width=1\linewidth]{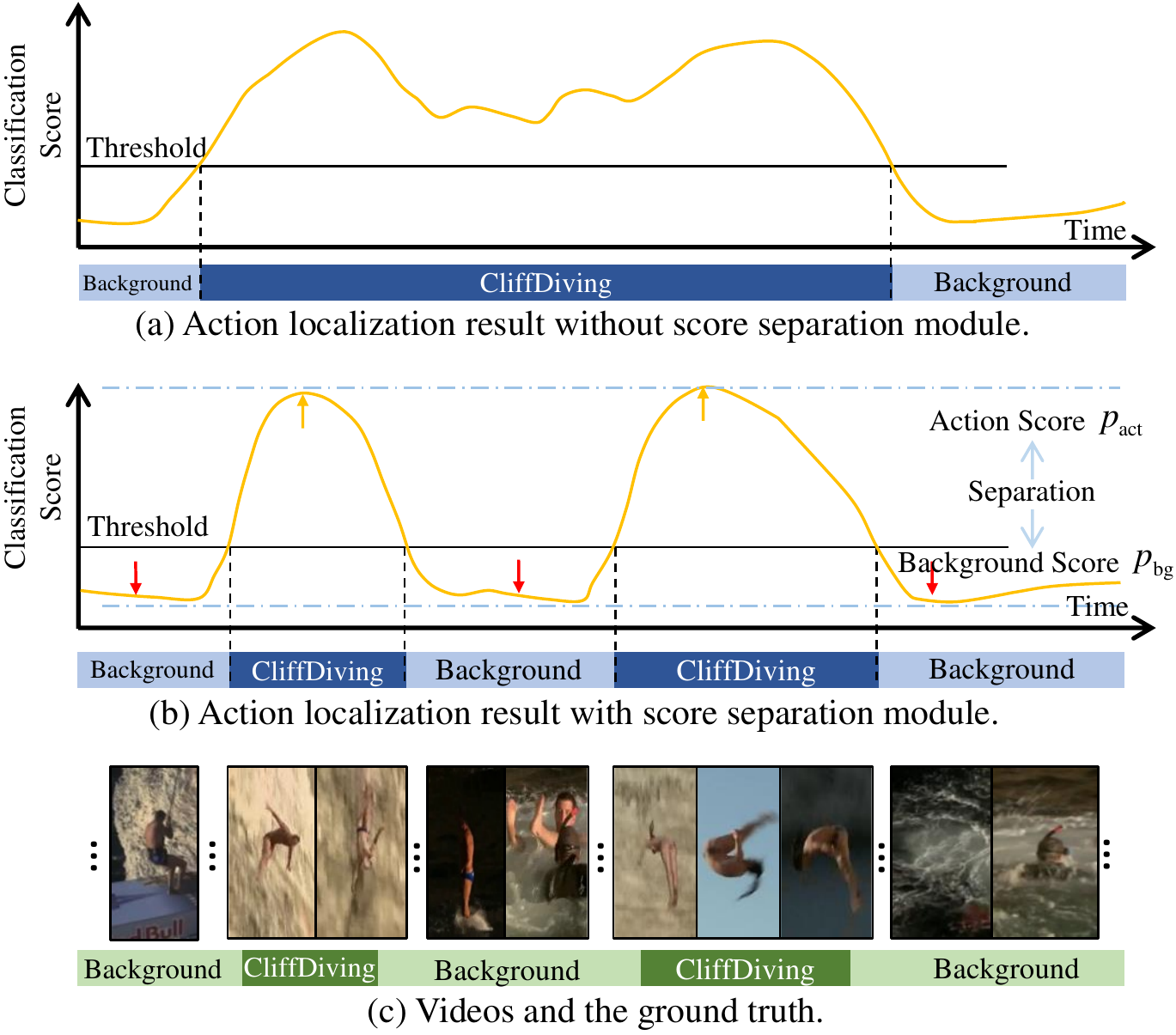}
	\caption{Class activation sequence before (a) and after (b)  employing the score separation module. We can find that the score separation module contributes to suppress the responses of background frames and enhance the responses of action frames, which is beneficial to separate adjacent action instances.}
	\label{fig:discrimination}
\end{figure}

\subsection{Score Separation Module}
\label{score-separation}

This section starts from the traditional weakly supervised action localization paradigm, and reveals that the top-$k$ aggregation procedure cannot explicitly influence the confusing frames. Then, we analyze the frame-level supervised classification in SF-Net \cite{ma2020sf}, and point out that it cannot thoroughly constrain action response. Afterwards, we propose the score separation module to utilize the position information within the click-level annotation and generate high-quality classification responses.

In the weakly supervised action localization paradigm, the top-$k$ aggregation procedure relies on the $k$ highest scores to predict the video-level classification score. In each training iteration, only the selected $k$ scores are influenced and optimized, while others would be ignored. Although the top-$k$ positions vary at the early training phase, a mature model would steadily select similar top-$k$ positions at the later training phase. Consequently, as shown in Fig. \ref{fig:discrimination}(a), the predicted classification score confidently shows high responses for action frames, but cautiously shows low responses for background frames. As long as scores of these confusing frames are lower than top-$k$ scores, they would not influence the video-level classification score. Thus, the responses of the action frames and the confusing background frames cannot be clearly separable, leading to imprecise prediction of the subsequent thresholding-based temporal action localization process.

To separate responses of actions and backgrounds, SF-Net \cite{ma2020sf} makes the action-click annotation and performs frame-level supervised classification. There are also other similar choices, such as performing the binary classification to learn actionness \cite{ma2020sf} and employing supervision on attention weights \cite{lee2020background}. However, based on our investigation (see section \ref{mining-position-information}), multiple variants of performing frame-level supervised classification are prone to obtain coessential information and cannot additively improve the action localization performance. Essentially, the frame-level cross-entropy loss can encourage the response of background class to be higher than the response of other classes, which implicitly suppresses the responses of all action classes. However, considering a video containing actions from $c^{th}$ class, the background frame-level cross-entropy loss cannot explicitly enforce the response of the $c^{th}$ class to be as low as possible on background frames, \eg lower than responses of all other action classes.

In this work, we explicitly constrain the response at background positions by using the score separation module, as shown in Fig. \ref{fig:discrimination}(b). In particular, given a video containing actions from $c^{th}$ category, we regard top-$k$ highest scores as the potential actions and calculate the mean score $p_{\rm act}$ via:
\begin{equation}
p_{\rm act}=\frac{1}{k}\sum_{\forall \hat{b}_{t}=0} s^{c}_{t},
\end{equation}
where $\hat{b}_{t}=0$ indicates top-$k$ action frames, whose total number is $k$. Similarly, given $N_{\rm frame}$ annotated background frames, the mean score $p_{\rm bg}$ is defined as:
\begin{equation}
p_{\rm bg}=\frac{1}{N_{\rm frame}} \sum_{\forall b_{t}=1} s^{c}_{t}.
\end{equation}

To enlarge the relative difference between mean action score $p_{\rm act}$ and mean background score $p_{\rm bg}$, we perform \textit{Softmax} normalization over $p_{act}$ and $p_{bg}$ as follows:
\begin{equation}
    \hat{p}_{act} = \frac{{\rm e}^{p_{act}}}{{\rm e}^{p_{act}} + {\rm e}^{p_{bg}}}, \ \ \hat{p}_{bg} = \frac{{\rm e}^{p_{bg}}}{e^{p_{act}} + {\rm e}^{p_{bg}}}.
\end{equation}
Afterwards, we guide $\hat{p}_{\rm act}$ to be one while $\hat{p}_{\rm bg}$ to be zero as follows:
\begin{equation}
\mathcal{L}_{\rm sep} = - {\rm log}\  \hat{p}_{\rm act} - {\rm log}\  (1-\hat{p}_{\rm bg}).
\end{equation}
The score separation loss $\mathcal{L}_{\rm sep}$ can guide the action response to be separated from the background response on the $c^{th}$ category.

\subsection{Affinity Module}
\label{affinity-module}

The affinity module is designed to explore the feature information within the background-click supervisions. Based on annotated background frames and pseudo action frames, we learn an embedding space for the input video. Then, considering a frame, we can measure its affinity with neighboring frames and obtain a frame-specific attention weight, namely local attention mask, which is injected into the convolutional calculation process. The frame-specific attention weight can guide the convolution process to dynamically attend to related neighbors, which generates more precise response.

In the affinity module, we first learn an embedding space to distinguish class-agnostic actions from backgrounds. Given input features, we use a temporal convolutional layer to learn embedding for each frame, \ie $\mathbf{E}=[\mathbf{e}_{1},\mathbf{e}_{2},...,\mathbf{e}_{T}]$, where $\mathbf{e}_{t} \in \mathbb{R}^{D_{\rm emb}}$ is a $D_{\rm emb}$-dimension vector. Each embedding vector is $L2-$normalized. Specifically, we use the cosine similarity to measure the affinity between two embeddings $\mathbf{e}_{u}$ and $\mathbf{e}_{v}$:
\begin{equation}
	\mathcal{A}(\mathbf{e}_{u},\mathbf{e}_{v})=\frac{\mathbf{e}^{T}_{u} \cdot \mathbf{e}_{v}}{\left\|\mathbf{e}_{u}\right\|_{2} \cdot \left\|\mathbf{e}_{v}\right\|_{2}}
	\label{equ-cosine-similarity}.
\end{equation}

Based on the annotated background frames and potential action frames, we can calculate the affinity loss $\mathcal{L}_{\rm aff}$ from three terms, \ie between two background frames, between two action frames and between the action-background pair. Particularly, we employ the online hard example mining strategy \cite{shrivastava2016training} to constrain the training frame pair. For the first term, embedding vectors of two background frames should be similar to each other, and the loss $\mathcal{L}^{\rm bg}_{\rm aff}$ can be formulated as:
\begin{equation}
\mathcal{L}^{\rm bg}_{\rm aff}=\max \limits_{\forall b_{u}=1, b_{v}=1, u\neq v} \lfloor \tau_{\rm same}-\mathcal{A}(\mathbf{e}_{u},\mathbf{e}_{v}) \rfloor_{+},
\end{equation}
where $\lfloor \cdot \rfloor_{+}$ denotes clipping bellowing at zero, $\tau_{\rm same}$ is the similarity threshold between frames from the same category. Specifically, we constrain the similarity between the two most dissimilar background frames should be larger than $\tau_{\rm same}$. Likewise, embedding vectors for action frames should be similar to each other as well:
\begin{equation}
\mathcal{L}^{\rm act}_{\rm aff}=\max \limits_{\forall \hat{b}_{u}=0, \hat{b}_{v}=0, u\neq v} \lfloor \tau_{\rm same}-\mathcal{A}(\mathbf{e}_{u},\mathbf{e}_{v}) \rfloor_{+}.
\end{equation}

\begin{figure}[t]
	\centering
	\graphicspath{{figure/}}
	\includegraphics[width=1\linewidth]{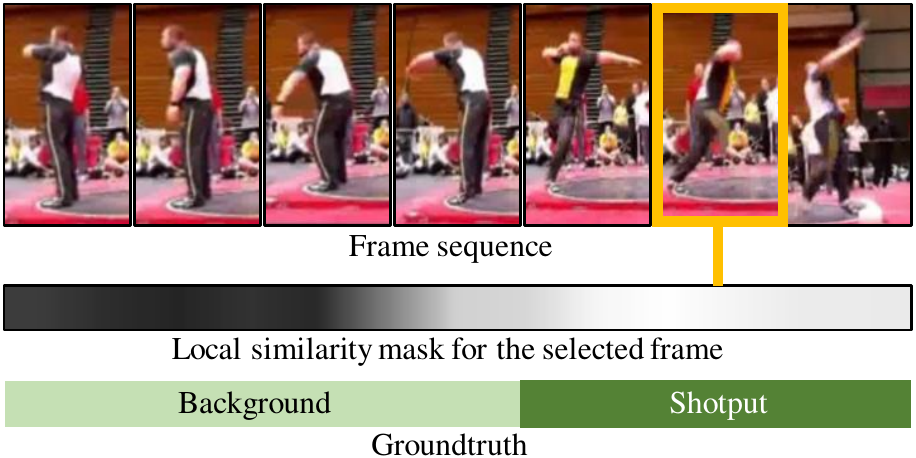}
	\caption{Visualization of the local similarity mask. Given a video containing the \textit{shotput} action, we select an action frame (shown in orange), and calculate similarities between the selected frame and its local neighbors. The generated local similarity mask exhibits high response for action frames and low response for background frames.}
	\label{fig:similarity}
\end{figure}

In contrast to $\mathcal{L}^{\rm bg}_{\rm aff}$ and $\mathcal{L}^{\rm act}_{\rm aff}$, embedding vectors of background frames should differ from embedding vectors of action frames. This can be formulated as:
\begin{equation}
\mathcal{L}^{\rm diff}_{\rm aff}=\max \limits_{\forall b_{u}=1, \hat{b}_{v}=0} \lfloor \mathcal{A}(\mathbf{e}_{u},\mathbf{e}_{v}) - \tau_{\rm diff} \rfloor_{+}
\end{equation}
where $\tau_{\rm diff}$ is the threshold to constrain the similarity between actions and backgrounds. The affinity loss jointly considers the above three terms and can be calculated as:
\begin{equation}
\mathcal{L}_{\rm aff} = \mathcal{L}^{\rm bg}_{\rm aff} + \mathcal{L}^{\rm act}_{\rm aff} + \mathcal{L}^{\rm diff}_{\rm aff}
\end{equation}

When high-quality embedding vectors are obtained, we can measure the cosine similarity between a frame and its local neighbors. As shown in Fig. \ref{fig:similarity}\footnote{Because some videos are shot from a distant perspective, the undergoing action would be small and hard to recognize when exhibited in original video frames. We follow previous works \cite{zhao2020temporal,shi2020weakly} to exhibit the undergoing action with cropped frames.}, the embedding vectors can distinguish action frames from background frames, and uniformly highlight coessential frames when given the reference frame.

Consider a video feature $\mathbf{X} \in \mathbb{R}^{D_{in} \times T}$ whose dimension is $D_{in}$ and temporal length is $T$, the temporal convolutional operation learns the convolutional kernel $\mathcal{H} \in \mathbb{R}^{h \times D_{in} \times D_{out}}$ to dispose of video feature $\mathbf{X}$, where $h$ is the size of the temporal convolutional kernel and $D_{out}$ is the dimension of the output feature. For simplicity, we only consider the $m^{th}$ channel of the output feature and use the convolutional kernel $\mathcal{H}^{m} \in \mathbb{R}^{h \times D_{in}}$. Then, the vanilla temporal convolutional operation for the $t^{th}$ feature can be formulated as:
\begin{equation}
\overline{f}^{m}_{t} = \sum_{i=0}^{h-1} \mathcal{H}^{m}[i, :] \cdot \mathbf{X}[:, t- \lfloor \frac{h}{2} \rfloor +i],
\end{equation}
where $[\cdot]$ means indexing data from the matrix, $\overline{f}^{m}_{t}$ is the value in the $m^{th}$ channel of the output feature vector, $(\cdot)$ indicates inner product, and $\lfloor \cdot \rfloor$ means round down.

Given a video, we calculate the local similarity for each temporal position and obtain affinity matrix $\mathbf{a} \in \mathbb{R}^{h \times T}$, where $\mathbf{a}[:, t]$ indicates the affinity between the $t^{th}$ feature vector and its $h$ neighbors. In contrast to vanilla convolution, we employ the affinity weight to modulate neighboring features of the $t^{th}$ position before performing temporal convolution:
\begin{equation}
    \overline{\mathbf{X}}[:, t- \lfloor \frac{h}{2} \rfloor +i] = \mathbf{X}[:, t- \lfloor \frac{h}{2} \rfloor +i] \times \mathbf{a}[i, t], i \in [0, ..., h-1],
\end{equation}
where $\overline{\mathbf{X}}$ is the modulated feature. Then, we perform temporal convolution on the $t^{th}$ position:
\begin{equation}
f^{m}_{t} = \sum_{i=0}^{h-1} \mathcal{H}^{m}[i, :] \cdot \overline{\mathbf{X}}[:, t- \lfloor \frac{h}{2} \rfloor +i].
\end{equation}
\vspace{-0.3cm}

In traditional methods, all temporal frames are tackled by the sharing convolutional kernel. In contrast, the frame-specific affinity weight guides the convolution to make the frame-specific calculation. Based on background frames and potential action frames, the affinity module adequately mines feature information. The affinity weights and frame-specific temporal convolution help distinguish actions from confusing backgrounds, which is beneficial to separate two closely adjacent actions. Although the affinity module contains three loss terms (\ie $\mathcal{L}_{\rm aff}^{\rm bg}$, $\mathcal{L}_{\rm aff}^{\rm act}$, and $\mathcal{L}_{\rm aff}^{\rm diff}$), each term can be effectively calculated via the similarity measurement based on matrix multiplication.

\subsection{Inference}
\label{method-action-localization}
In inference, we forward a testing video through the learned network and obtain the class activation sequence $\mathbf{S}$. Then, the top-$k$ aggregation procedure predicts the video-level classification score $\mathbf{s}_{v}$. Among $C$ candidate categories, we discard categories whose video-level classification score is lower than threshold $\tau_{\rm cls}$. Next, we take the class activation sequence for the confident categories and regard consecutive frames with high scores as action instances, obtaining the start time $t_{i}^{s}$ and the end time $t_{i}^{e}$. Afterwards, the confidence score $p_{i}$ for this predicted action instance is determined via outer-inner-contrastive strategy \cite{shou2018autoloc}.

\vspace{-0.3cm}
\section{Experiments}
\label{sec-experiments}
In this section, we carry experiments to evaluate and analyze the proposed BackTAL method. We start from experimental setups in section \ref{sec-experimental-setups}. Then, section \ref{sec-click-annotation} presents the annotation process for background-click information. Next, we compare BackTAL with recent start-of-the-art methods on three benchmark datasets and verify the superior performance of BackTAL in section \ref{sec-cmp-exps}. Afterwards, section \ref{section-ablation} carries ablation studies to analyze the superiority of background-click supervision, the effectiveness of each module, and studies the influence of parameters. Additionally, we depict qualitative analysis in section \ref{sec-qual-exps}.

\vspace{-0.3cm}
\subsection{Experimental Setups}
\label{sec-experimental-setups}
\textbf{Benchmark Datasets.}
We evaluate the efficacy of BackTAL on three benchmarks, including THUMOS14 \cite{THUMOS14}, ActivityNet v1.2 \cite{caba2015activitynet}, and HACS \cite{zhao2019hacs}. In THUMOS14, the training set consists of 2765 trimmed videos, while the validation set and test set consist of 200 and 213 untrimmed videos, respectively. As a common practice in the literature \cite{wang2017untrimmednets, nguyen2018weakly, lee2020background}, we employ the validation set in the training phase and evaluate the performance on the test set, where videos are from 20 classes. The main challenge in THUMOS14 is dramatic variation of action instances' duration. Specifically, a short action instance only lasts tenths of a second, while a long action instance can last hundreds of seconds \cite{xu2017r,chao2018rethinking,xu2019two}. ActivityNet v1.2 \cite{caba2015activitynet} includes 9682 videos from 100 classes, which are divided into training, validation, and testing subsets via the ratio 2:1:1. Challenges in ActivityNet v1.2 usually lie on numerous action categories, large intra-class variations, \etc. In addition to these two commonly used datasets, we notice a recently proposed dataset HACS \cite{zhao2019hacs}. It contains 50 thousand videos spanning 200 classes, where training set, validation set, and testing set consist of 38 thousand, 6 thousand, and 6 thousand videos, respectively. Compared with existing benchmarks, HACS contains large-scale videos and action instances, serving as a more realistic and challenging benchmark. In addition, we follow SF-Net \cite{ma2020sf} and evaluate the performance of BackTAL on BEOID dataset \cite{damen2014you}. BEOID consists of 58 videos, 742 action instances, coming from 34 action categories. We make the background-click annotation for videos on BEOID dataset.

\textbf{Evaluation Metric.}
Mean average precision (mAP) under different thresholds \cite{THUMOS14, caba2015activitynet} is used to evaluate the performance. On THUMOS14, we report mAP under thresholds tIoU=\{0.3,0.4,0.5,0.6,0.7\}, and follow previous works \cite{liu2019weakly, yu2019temporal,luo2020weakly} to focus on mAP@tIoU0.5. Besides, considering some methods may exhibit superiority on low or high tIoU threshold, we report average mAP under thresholds tIoU=\{0.3,0.4,0.5,0.6,0.7\} to perform a holistic comparison, as Liu \etal \cite{liu2019completeness} have tried. The evaluation on ActivityNet and HACS employ the average mAP under ten uniformly distributed thresholds from tIoU=0.5 to tIoU=0.95, \ie [0.5:0.05:0.95]. On BEOID \cite{damen2014you}, we follow SF-Net \cite{ma2020sf} to report mAP under threshold [0.1:0.1:0.7] as well as the average value of these seven mAPs.

\textbf{Baseline Method.} We follow a recent work BaS-Net \cite{lee2020background} to build our baseline method, principally considering its simplicity. The network utilizes three temporal convolutional layers to perform classification for video frames and to generate the class activation sequence. For each video, this network is used twice. The first time disposes of basic video features and the second time tackles filtered video features. We make one simplification over the official implementation of BaS-Net and improve the performance from 27.0 to 28.6, under the metric mAP(\%)@tIoU0.5. Specifically, BaS-Net \cite{lee2020background} performs data augmentation via randomly selecting a part of the video, but we scale video features to fixed temporal length and use complete features via linear interpolation, following \cite{long2019gaussian, lin2019bmn}. A potential reason for the improvement is that, selecting a part of video is faced with a certain risk to drop some action instances or to cut one complete action instance into a part, which may confuse the learning algorithm.

\textbf{Implementation Details.}
Following previous works \cite{nguyen2018weakly, liu2019completeness, shi2020weakly} , we use the I3D \cite{carreira2017quo} model pre-trained on the Kinetics-400 \cite{carreira2017quo} dataset to extract both RGB and optical flow features. As for scaled feature sequence, the temporal length $T$ for THUMOS14, ActivityNet v1.2, and HACS is 750, 100 and 200, respectively. The top-$k$ aggregation procedure selects $k=\lfloor \frac{1}{8} \times T \rfloor$ highest scores on each dataset, where $\lfloor \cdot \rfloor$ means floor down.

\begin{figure}[thbp]
	\graphicspath{{figure/}}
	\centering
	\includegraphics[width=1\linewidth]{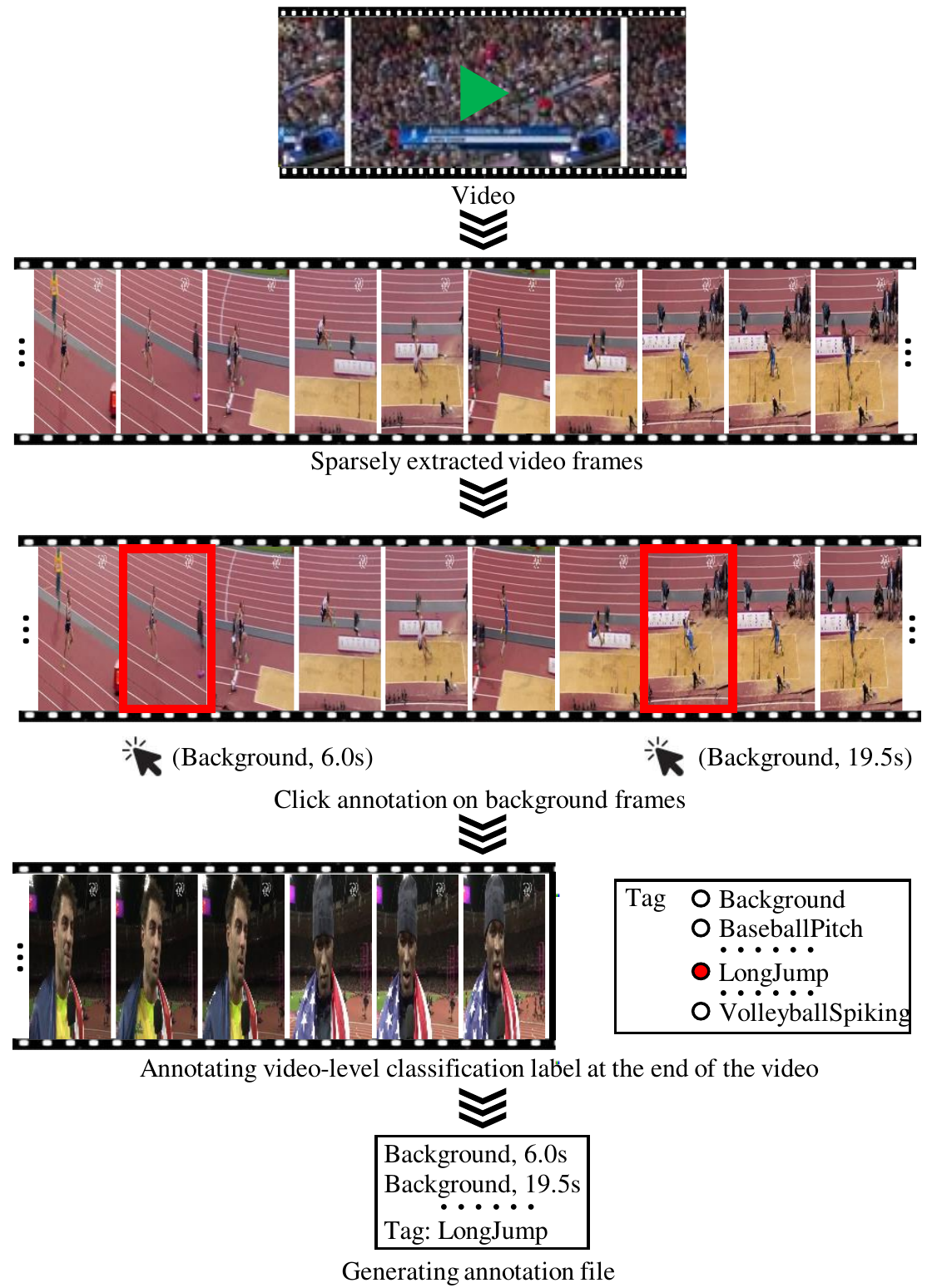}
	\vspace{-0.4cm}
	\caption{Processes to annotate the background-click information, illustrated with a video containing the action \textit{LongJump}. Firstly, we sparsely extract frames from the video with 2\textit{fps}. Then, when meeting a background segment, the annotator randomly clicks a frame and annotates it as background. Afterwards, the video-level classification label is recorded at the end of the video. Finally, the annotation file can be generated for the complete video.}
	\label{ann-proc}
	\vspace{-0.4cm}
\end{figure}

The proposed BackTAL is implemented on PyTorch 1.5 \cite{paszke2019pytorch} and optimized via the Adam algorithm. We use batch size 16, learning rate $1 \times 10^{-4}$ and weight decay $5 \times 10^{-4}$. We train 100, 25 and 8 epochs for THUMOS14, ActivityNet v1.2 and HACS, respectively. We set embedding dimension as $\mathbb{R}^{D_{\rm emb}}=32$. For fair comparison, we follow BaS-Net \cite{lee2020background} to set hyper-parameters. Specifically, we adopt the same inference paradigm with BaS-Net and set $\tau_{\rm cls}$=0.25. We employ the gird search strategy to empirically determine the proper values for hyper-parameters. Specifically, the balancing coefficients are set as $\lambda=$1, $\beta=$0.8. In affinity loss, we set $\tau_{\rm same}=$0.5, $\tau_{\rm diff}=$0.1. The influence of these hyper-parameters are discussed via ablation experiments in section \ref{section-ablation}.

\vspace{-0.4cm}
\subsection{Background-Click Annotation}
\label{sec-click-annotation}
Before conducting experiments, we make the background-click annotation on THUMOS14 \cite{THUMOS14}. To start with, we train three annotators with a few actions and backgrounds to make them familiar with each action category. Then, annotators are requested to randomly annotate a background frame once they see a new background segment. Employing the annotation tool provided by Tang \etal \cite{tang2020comprehensive}, annotators can quickly skim action frames and make efficient annotations. Fig. \ref{ann-proc} exhibits the detailed annotation process. Specifically, as the sparse extraction can reduce frame number and speed up the annotation process, we extract frames with the frame rate of 2\textit{fps}. In a video, we click background frames and only record the video-level classification label at the end of the video. As a result, the annotation process is efficient. On average, it takes 48 seconds to annotate a one-minute video.\footnote{In addition, we explore the cost to annotate both action clicks and background clicks for a one-minute video, and spend 53s after sparse frame extraction.}

\begin{table*}[htbp]
	\centering
	\caption{Comparison experiments on THUMOS14 dataset. We compare BackTAL with three fully-supervised methods (instance-level supervision), recent weakly-supervised methods (video-level supervision) and weakly-supervised methods with extra informations (Video-level + $*$).}
	\begin{threeparttable}[t]
		\begin{tabular}{c|ccc|cccccc}
			\toprule
			\toprule
			\multirow{2}[4]{*}{Research} & \multirow{2}[4]{*}{Publication} & \multirow{2}[4]{*}{Feature} & \multirow{2}[4]{*}{Supervision} & \multicolumn{5}{c}{mAP@tIoU (\%)}     & \multicolumn{1}{l}{avg-mAP} \\
			\cmidrule{5-9}          &       &       &       & 0.3   & 0.4   & 0.5   & 0.6   & 0.7   & \multicolumn{1}{l}{(0.3:0.7)} \\
			\midrule
			\midrule
			R-C3D \cite{xu2017r} & ICCV 2017 & C3D   & Instance-level & 44.8  & 35.6  & \cellcolor[rgb]{ .906,  .902,  .902}28.9 & -     & -     & \cellcolor[rgb]{ .906,  .902,  .902}- \\
			BMN \cite{lin2019bmn} & ICCV 2019 & UNT   & Instance-level & \textbf{56.0} & 47.4  & \cellcolor[rgb]{ .906,  .902,  .902}38.8 & 29.7  & 20.5  & \cellcolor[rgb]{ .906,  .902,  .902}38.5 \\
			G-TAD \cite{xu2020g} & CVPR 2020 & UNT   & Instance-level & 54.5  & \textbf{47.6} & \cellcolor[rgb]{ .906,  .902,  .902}\textbf{40.2} & \textbf{30.8} & \textbf{23.4} & \cellcolor[rgb]{ .906,  .902,  .902}\textbf{39.3} \\
			\midrule
			\midrule
			TSRNet \cite{zhang2019learning} & AAAI 2019 & ReSNet-101 & Video-level & 38.3  & 28.1  & \cellcolor[rgb]{ .906,  .902,  .902}18.6 & 11.0  & 5.6   & \cellcolor[rgb]{ .906,  .902,  .902}20.3 \\
			Xu \etal  \cite{xu2019segregated} & AAAI 2019 & I3D   & Video-level & 48.7  & 34.7  & \cellcolor[rgb]{ .906,  .902,  .902}23.0 & -     & -     & \cellcolor[rgb]{ .906,  .902,  .902}- \\
			CMCS \cite{liu2019completeness} & CVPR 2019 & I3D   & Video-level & 41.2  & 32.1  & \cellcolor[rgb]{ .906,  .902,  .902}23.1 & 15.0  & 7.0   & \cellcolor[rgb]{ .906,  .902,  .902}23.7 \\
			Yu \etal \cite{yu2019temporal} & ICCV 2019 & I3D   & Video-level & 39.5  & 31.9  & \cellcolor[rgb]{ .906,  .902,  .902}24.5 & 13.8  & 7.1   & \cellcolor[rgb]{ .906,  .902,  .902}23.4 \\
			BaS-Net \cite{lee2020background} & AAAI 2020 & I3D   & Video-level & 44.6  & 36.0  & \cellcolor[rgb]{ .906,  .902,  .902}27.0 & 18.6  & 10.4  & \cellcolor[rgb]{ .906,  .902,  .902}27.3 \\
			TSCN \cite{zhai2020two} & ECCV 2020 & I3D   & Video-level & 47.8  & 37.7  & \cellcolor[rgb]{ .906,  .902,  .902}28.7 & 19.4  & 10.2  & \cellcolor[rgb]{ .906,  .902,  .902}28.8 \\
			DGAM \cite{shi2020weakly} & CVPR 2020 & I3D   & Video-level & 46.8  & 38.2  & \cellcolor[rgb]{ .906,  .902,  .902}28.8 & 19.8  & 11.4  & \cellcolor[rgb]{ .906,  .902,  .902}29.0 \\
			Liu \etal \cite{liu2021weakly} & AAAI 2021 & I3D   & Video-level & 50.8  & 41.7  & \cellcolor[rgb]{ .906,  .902,  .902}29.6 & 20.1  & 10.7  & \cellcolor[rgb]{ .906,  .902,  .902}30.6 \\
			Gong \etal \cite{gong2020learning} & CVPR 2020 & I3D   & Video-level & 46.9  & 38.9  & \cellcolor[rgb]{ .906,  .902,  .902}30.1 & 19.8  & 10.4  & \cellcolor[rgb]{ .906,  .902,  .902}29.2 \\
			A2CL-PT \cite{min2020adversarial} & ECCV 2020 & I3D   & Video-level & 48.1  & 39.0  & \cellcolor[rgb]{ .906,  .902,  .902}30.1 & 19.2  & 10.6  & \cellcolor[rgb]{ .906,  .902,  .902}29.4 \\
			EM-MIL \cite{luo2020weakly} & ECCV 2020 & I3D   & Video-level & 45.5  & 36.8  & \cellcolor[rgb]{ .906,  .902,  .902}30.5 & 22.7  & 16.4  & \cellcolor[rgb]{ .906,  .902,  .902}30.4 \\
			ACSNet \cite{liu2021acsnet} & AAAI 2021 & I3D   & Video-level & 51.4  & 42.7  & \cellcolor[rgb]{ .906,  .902,  .902}32.4 & 22.0  & 11.7  & \cellcolor[rgb]{ .906,  .902,  .902}32.0 \\
			ACM-BANet \cite{moniruzzaman2020action} & ACM MM 2020 & I3D   & Video-level & 48.9  & 40.9  & \cellcolor[rgb]{ .906,  .902,  .902}32.3 & 21.9  & 13.5  & \cellcolor[rgb]{ .906,  .902,  .902}31.5 \\
			HAM-Net \cite{islam2021a} & AAAI 2021 & I3D   & Video-level & 52.2  & 43.1  & \cellcolor[rgb]{ .906,  .902,  .902}32.6 & 21.9  & 12.5  & \cellcolor[rgb]{ .906,  .902,  .902}32.5 \\
			Lee \etal \cite{lee2021weakly} & AAAI 2021 & I3D   & Video-level & 52.3  & 43.4  & \cellcolor[rgb]{ .906,  .902,  .902}33.7 & 22.9  & 12.1  & \cellcolor[rgb]{ .906,  .902,  .902}32.9 \\
			\midrule
			3C-Net \cite{narayan20193c} & ICCV 2019 & I3D   & Video-level + action count & 44.2  & 34.1  & \cellcolor[rgb]{ .906,  .902,  .902}26.6 & -     & 8.1   & \cellcolor[rgb]{ .906,  .902,  .902}- \\
			Nguyen \etal \cite{nguyen2019weakly} & ICCV 2019 & I3D   & Video-level + microvideos & 49.1  & 38.4  & \cellcolor[rgb]{ .906,  .902,  .902}27.5 & 17.3  & 8.6   & \cellcolor[rgb]{ .906,  .902,  .902}28.2 \\
			ActionBytes \cite{jain2020actionbytes} & CVPR 2020 & I3D   & Video-level + Kinetics val & 43.0  & 37.5  & \cellcolor[rgb]{ .906,  .902,  .902}29.0 & -     & 9.5   & \cellcolor[rgb]{ .906,  .902,  .902}- \\
			SF-Net \cite{ma2020sf} & ECCV 2020 & I3D   & Video-level + click-level & 52.8  & 42.2  & \cellcolor[rgb]{ .906,  .902,  .902}30.5 & 20.6  & 12.0  & \cellcolor[rgb]{ .906,  .902,  .902}31.6 \\
			BackTAL & -     & I3D   & Video-level + click-level & \textbf{54.4} & \textbf{45.5} & \cellcolor[rgb]{ .906,  .902,  .902}\textbf{36.3} & \textbf{26.2} & \textbf{14.8} & \cellcolor[rgb]{ .906,  .902,  .902}\textbf{35.4} \\
			\bottomrule
			\bottomrule
		\end{tabular}%
		\begin{tablenotes}
			\item \textit{As for feature extract, most works utilize I3D model \cite{carreira2017quo}. BMN \cite{lin2019bmn}, G-TAD \cite{xu2020g} and CleanNet \cite{liu2019weakly} utilize UntrimmedNet model \cite{wang2017untrimmednets, wang2018temporal}. TSRNet utilizes ReSNet-101 \cite{he2016deep}.}
		\end{tablenotes}
	\end{threeparttable}
	\label{tab:cmp-thumos}%
	\vspace{-0.3cm}
\end{table*}%

\begin{figure}[htbp]
	\graphicspath{{Fig./}}
	\centering
	\includegraphics[width=1\linewidth]{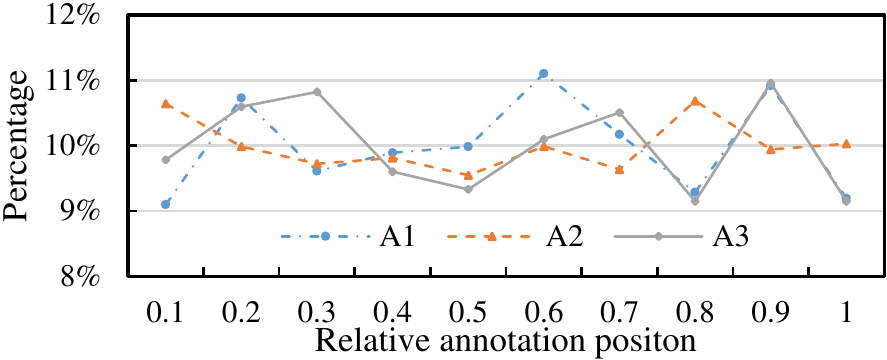}
	\caption{Statistics of background-click annotations on THUMOS14 dataset. The x-axis indicates the relative position for each annotation, while the y-axis indicates percentage of annotated frames. We can find that background-click annotations approximately exhibit the uniform distribution. ``A1", ``A2" and ``A3" indicate three different annotators.}
	\label{fig:ann_dis}
\end{figure}

We statistic the relative position of background-click annotations, with respect to the corresponding background segment. Considering a background segment starts at $t^{b}_{s}$ and ends at $t^{b}_{e}$, for a background-click annotation with timestamp $t^{b}$, the relative position can be calculated via $\frac{t^{b} - t^{b}_{s}}{t^{b}_{e} - t^{b}_{s}}$. As shown in Fig. \ref{fig:ann_dis}, annotation positions from three annotators approximately exhibit the uniform distribution. Potential reasons for uniform distribution include: first, the annotator randomly clicks a background frame within the background segment. Besides, because the background frame is easy to identify, the annotator hardly makes error. For experiments on THUMOS14, the performance of BackTAL is the average of three trials employing three different annotations. On THUMOS14, SF-Net \cite{ma2020sf} observes similar performances between human annotations and simulated annotations. Because the ActivityNet v1.2 contains dozens of times videos than THUMOS14, SF-Net \cite{ma2020sf} adopts a simulation strategy, \ie randomly annotating a frame within each action instance based on the ground truth. In this paper, we follow SF-Net and use the simulated annotations on large-scale datasets ActivityNet v1.2 \cite{caba2015activitynet} and HACS \cite{zhao2019hacs}.

\begin{table*}[htbp]
	\vspace{-0.2cm}
	\centering
	\caption{Comparison experiments on ActivityNet v1.2 dataset. We compare BackTAL with recent weakly-supervised methods (video-level supervision), and weakly-supervised methods with extra informations (video-level + $*$).}
	\setlength{\tabcolsep}{3.5pt}
	\begin{threeparttable}[t]
		\begin{tabular}{c|ccc|ccccccccccc}
			\toprule
			\toprule
			\multirow{2}[4]{*}{Research} & \multirow{2}[4]{*}{Publication} & \multirow{2}[4]{*}{Feature} & \multirow{2}[4]{*}{Supervision} & \multicolumn{10}{c}{mAP@tIoU (\%)}                                            & \multicolumn{1}{l}{avg-mAP} \\
			\cmidrule{5-14}          &       &       &       & 0.50  & 0.55  & 0.60  & 0.65  & 0.70  & 0.75  & 0.80  & 0.85  & 0.90  & 0.95  & 0.50:0.95 \\
			\midrule
			\midrule
			EM-MIL \cite{luo2020weakly} & ECCV 2020 & I3D   & Video-level & 37.4  & -     & -     & -     & 23.1  & -     & -     & -     & 2.0   & -     & \cellcolor[rgb]{ .906,  .902,  .902}20.3 \\
			CleanNet \cite{liu2019weakly} & ICCV 2019 & UNT   & Video-level & 37.1  & 33.4  & 29.9  & 26.7  & 23.4  & 20.3  & 17.2  & 13.9  & 9.2   & 5.0   & \cellcolor[rgb]{ .906,  .902,  .902}21.6 \\
			CMCS \cite{liu2019completeness} & CVPR 2019 & I3D   & Video-level & 36.8  & -     & -     & -     & -     & 22.0  & -     & -     & -     & 5.6   & \cellcolor[rgb]{ .906,  .902,  .902}22.4 \\
			TSCN \cite{zhai2020two} & ECCV 2020 & I3D   & Video-level & 37.6  & -     & -     & -     & -     & 23.7  & -     & -     & -     & \textbf{5.7} & \cellcolor[rgb]{ .906,  .902,  .902}23.6 \\
			BaSNet \cite{lee2020background} & AAAI 2020 & I3D   & Video-level & 38.5  & -     & -     & -     & -     & 24.2  & -     & -     & -     & 5.6   & \cellcolor[rgb]{ .906,  .902,  .902}24.3 \\
			DGAM \cite{shi2020weakly} & CVPR 2020 & I3D   & Video-level & 41.0  & 37.5  & 33.5  & 30.1  & 26.9  & 23.5  & 19.8  & 15.5  & 10.8  & 5.3   & \cellcolor[rgb]{ .906,  .902,  .902}24.4 \\
			Gong \etal \cite{gong2020learning} & CVPR 2020 & I3D   & Video-level & 40.0  & -     & -     & -     & -     & 25.0  & -     & -     & -     & 4.6   & \cellcolor[rgb]{ .906,  .902,  .902}24.6 \\
			HAM-Net \cite{islam2021a} & AAAI 2021 & I3D   & Video-level & 41.0  & -     & -     & -     & -     & 24.8  & -     & -     & -     & 5.3   & \cellcolor[rgb]{ .906,  .902,  .902}25.1 \\
			Liu \etal \cite{liu2021weakly} & AAAI 2021 & I3D   & Video-level & 39.2  & -     & -     & -     & -     & 25.6  & -     & -     & -     & 6.8   & \cellcolor[rgb]{ .906,  .902,  .902}25.5 \\
			Lee \etal \cite{lee2021weakly} & AAAI 2021 & I3D   & Video-level & 41.2  & -     & -     & -     & -     & 25.6  & -     & -     & -     & 6.0   & \cellcolor[rgb]{ .906,  .902,  .902}25.9 \\
			ACSNet \cite{liu2021acsnet} & AAAI 2021 & I3D   & Video-level & 40.1  & -     & -     & -     & -     & 26.1  & -     & -     & -     & 6.8   & \cellcolor[rgb]{ .906,  .902,  .902}26.0 \\
			\midrule
			\midrule
			3C-Net \cite{narayan20193c} & ICCV 2019 & I3D   & Video-level + action count & 37.2  & -     & -     & -     & 23.7  & -     & -     & -     & 9.2   & -     & \cellcolor[rgb]{ .906,  .902,  .902}21.7 \\
			SF-Net \cite{ma2020sf} & ECCV 2020 & I3D   & Video-level + click-level & 37.8  & -     & -     & -     & 24.6  & -     & -     & -     & 10.3  & -     & \cellcolor[rgb]{ .906,  .902,  .902}22.8 \\
			BackTAL & -     & I3D   & Video-level + click-level & \textbf{41.5} & \textbf{39.0} & \textbf{36.4} & \textbf{32.9} & \textbf{30.2} & \textbf{27.3} & \textbf{23.7} & \textbf{19.8} & \textbf{14.4} & 4.7   & \cellcolor[rgb]{ .906,  .902,  .902}\textbf{27.0} \\
			\bottomrule
			\bottomrule
		\end{tabular}%
	\end{threeparttable}
	\label{tab:cmp-anet12}%
\end{table*}%

\begin{table*}[t]
	\centering
	\caption{Comparison experiments on HACS dataset, in comparison with a fully-supervised SSN \cite{zhao2017temporal} and a weakly-supervised BaS-Net \cite{lee2020background}.}
	\begin{threeparttable}[t]
		\begin{tabular}{c|cc|cccccccccc|c}
			\toprule
			\toprule
			\multirow{2}[4]{*}{Research} & \multirow{2}[4]{*}{Publication} & \multirow{2}[4]{*}{Supervision} & \multicolumn{10}{c}{mAP@tIoU (\%)}                                            & \multicolumn{1}{l}{avg-mAP} \\
			\cmidrule{4-13}          &       &       & 0.50  & 0.55  & 0.60  & 0.65  & 0.70  & 0.75  & 0.80  & 0.85  & 0.90  & \multicolumn{1}{c}{0.95 } & 0.50:0.95 \\
			\midrule
			\midrule
			SSN \cite{zhao2017temporal} & ICCV 2017 & Instance-level & 28.8  & -     & -     & -     & -     & 18.8  & -     & -     & -     & 5.3   & \cellcolor[rgb]{ .906,  .902,  .902}19.0 \\
			\midrule
			\midrule
			BaS-Net \cite{lee2020background} & AAAI 2020 & Video-level & 30.6  & 27.7  & 25.1  & 22.6  & 20.0  & 17.4  & 14.8  & 12.0  & 9.2   & \textbf{5.7} & \cellcolor[rgb]{ .906,  .902,  .902}18.5 \\
			\midrule
			BackTAL & -     & Video-level + click-level & \textbf{31.5} & \textbf{29.1} & \textbf{26.8} & \textbf{24.5} & \textbf{22.0} & \textbf{19.5} & \textbf{17.0} & \textbf{14.2} & \textbf{10.8} & 4.7   & \cellcolor[rgb]{ .906,  .902,  .902}\textbf{20.0} \\
			\bottomrule
			\bottomrule
		\end{tabular}%
		\begin{tablenotes}
			\item \textit{Results of SSN \cite{zhao2017temporal} are taken from \cite{zhao2019hacs}. Results of BaS-Net \cite{lee2020background} are from our implementation.}
		\end{tablenotes}
	\end{threeparttable}
	\label{tab:cmp-hacs}%
\end{table*}%

\subsection{Comparison with State-of-the-Art Methods}
\label{sec-cmp-exps}
\textbf{THUMOS14.} 
Table \ref{tab:cmp-thumos} compares BackTAL with recent state-of-the-art methods on THUMOS14 dataset. As this paper focuses on weakly supervised temporal action localization, we only list three representative supervised methods \cite{xu2017r, lin2019bmn, xu2020g} to indicate the progress under the supervised paradigm. For weakly supervised paradigm, we distinguish methods only using the video-level classification label from methods that use extra information. In Table \ref{tab:cmp-thumos}, the most similar competitor to the proposed BackTAL is SF-Net \cite{ma2020sf}, where SF-Net employs action-click supervision and BackTAL employs background-click supervision. Under similar annotation cost, BackTAL exhibits 5.8 mAP improvements over SF-Net under tIoU threshold 0.5, demonstrating background-click supervision is more effective. Besides, with the rapid development of weakly supervised methods, some recent works \cite{liu2021acsnet, moniruzzaman2020action, islam2021a, lee2021weakly} achieve superior performance than weakly supervised methods employing extra information. However, the proposed BackTAL performs 2.6 mAP higher than current well-performed method \cite{lee2021weakly} under tIoU threshold 0.5. Moreover, in comparison with supervised methods, BackTAL can exceed a classical method \cite{xu2017r} but still shows obvious performance gap with recent supervised methods. This indicates the weakly supervised methods should be persistently developed.

\textbf{ActivityNet v1.2.} 
Table \ref{tab:cmp-anet12} reports the performance of BackTAL and current state-of-the-art methods on ActivityNet v1.2 benchmark. ActivityNet v1.2 possesses different characteristics with THUMOS14, \eg a large percentage of action instances are extremely long, dramatic variations within an action instance. The previous counterpart SF-Net \cite{ma2020sf} principally performs supervised classification on action-click annotated frames, which is effective to learn similar patterns within neighboring frames but is insufficient to propagate information over long-range interval. As a result, SF-Net exhibits inferior performance to some weakly supervised methods \cite{islam2021a,liu2021weakly,lee2021weakly,liu2021acsnet}. In contrast, the proposed BackTAL converts valuable click-level supervision to background segments and discovers action instances through video-level classification process, \ie the top-$k$ aggregation process. As shown in Table \ref{tab:cmp-anet12}, BackTAL performs favorable over recent weakly supervised methods, under metric average mAP. Under ten different thresholds, BackTAL achieves high performance on nine thresholds. As tIoU threshold 0.95 is a strict criteria, a potential reason is that there is a trade-off between the performance on the holistic dataset and the precise boundary localization on some action instances. BackTAL focuses on the holistic performance and achieves high average mAP.

\textbf{HACS.} In addition to two traditional benchmarks, we make an early attempt and verify the effectiveness of weakly supervised temporal action localization on the large-scale HACS dataset, shown in Table \ref{tab:cmp-hacs}. SSN \cite{zhao2017temporal} is a classical fully supervised temporal action localization method. It models action structure with a pyramid architecture, and employs the activity classifier and completeness classifier to predict action category and completeness score, respectively. As a weakly supervised method, BaS-Net \cite{lee2020background} shows inferior performance than SSN under the metric average mAP.
\begin{table}[thbp]
  \centering
  \setlength{\tabcolsep}{4.0pt}
  \caption{Comparison experiments on BEOID dataset, measured by mAP under different tIoU threshold.}
    \begin{tabular}{c|cccccccc}
    \toprule
    \toprule
    \multirow{2}[4]{*}{Research} & \multicolumn{7}{c}{mAP@tIoU (\%)}                     & avg-mAP \\
\cmidrule{2-8}          & 0.1   & 0.2   & 0.3   & 0.4   & 0.5   & 0.6   & 0.7   & (0.1:0.7) \\
    \midrule
    \midrule
    SF-Net \cite{ma2020sf} & \textbf{62.9} & -     & 40.6  & -     & 16.7  & -     & 3.5   & 30.1 \\
    BackTAL & 60.1  & 49.5  & \textbf{40.9} & 30.8  & \textbf{21.2} & 14.0  & \textbf{11.0} & \textbf{32.5} \\
    \bottomrule
    \bottomrule
    \end{tabular}%
  \label{expBEOID}%
\end{table}%
In contrast, the proposed BackTAL exceeds SSN under nine over ten different tIoU thresholds as well as the average mAP. Considering HACS \cite{zhao2019hacs} is a large-scale realistic dataset, this experiment reveals the promising prospect of the background-click supervision.

\textbf{BEOID.} Table \ref{expBEOID} reports the performance comparison between SF-Net \cite{ma2020sf} and the proposed BackTAL. SF-Net utilizes the action-click supervision and achieves 30.1 mAP, under the metric average mAP. BackTAL employs the background-click supervision, achieves 32.5 mAP, and exhibits 2.4 mAP improvements over SF-Net. In Table \ref{expBEOID}, it can be noticed that SF-Net performs well under low tIoU threshold (\eg 0.1). One potential reason is that action-click supervision contributes to discovering action instances, but can only generate coarse temporal boundaries. In general, BackTAL performs well under other tIoU thresholds and average mAP, which demonstrates the effectiveness of the proposed background-click supervision.

\begin{figure}[thbp]
	\centering
	\includegraphics[width=1\linewidth]{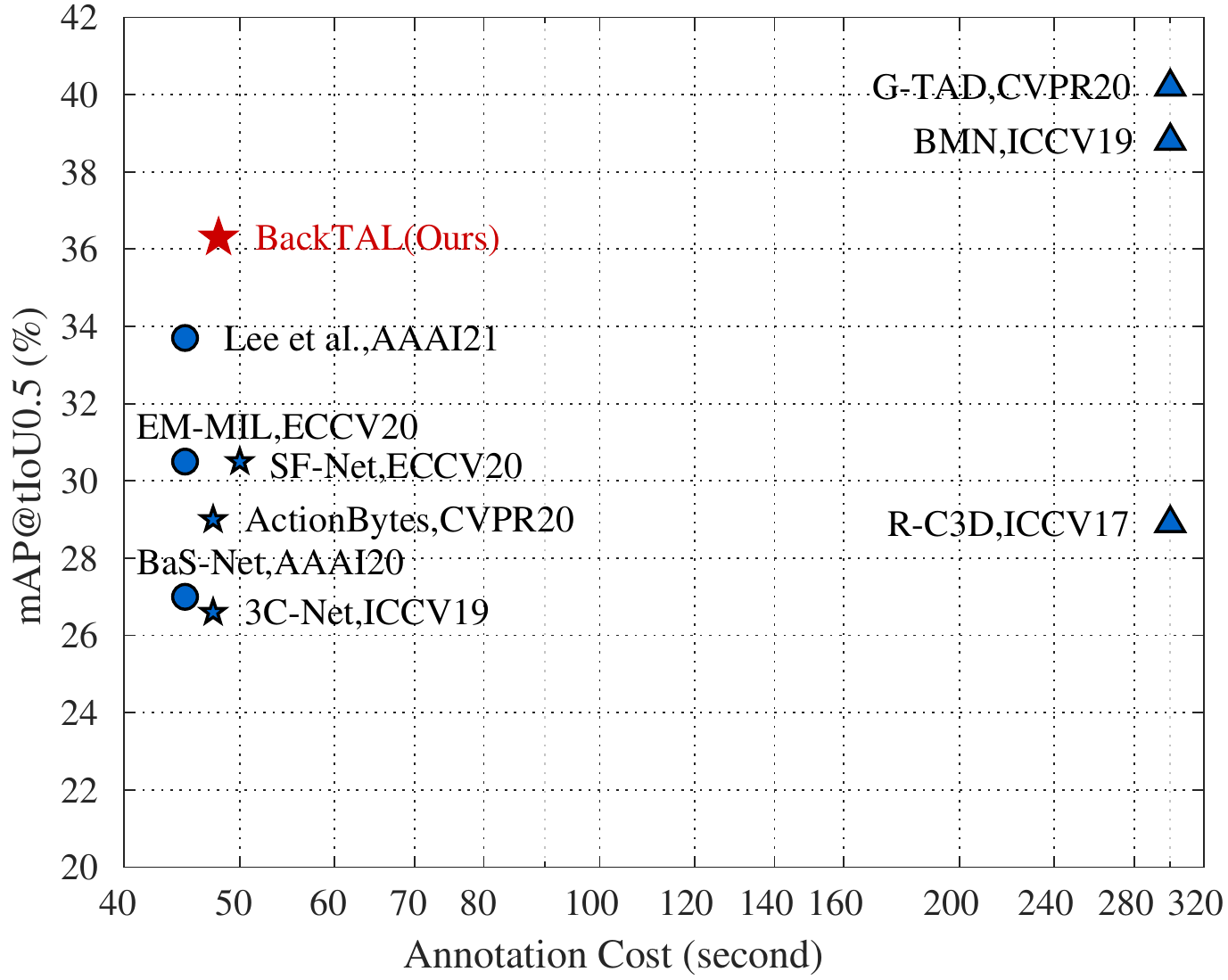}
	\caption{Trade-off between annotation cost and action localization performance on THUMOS14 dataset. We compare BackTAL with recent methods that employ weak supervision, weak supervision with extra information, and full supervision. Annotation costs for weakly supervised methods \cite{lee2020background, ma2020sf, lee2021weakly} and fully supervised methods \cite{xu2017r, lin2019bmn, xu2020g} are taken from \cite{ma2020sf}.} The x-axis is the log-scale.
	\label{fig-trade-off}
\end{figure}

\begin{table}[t]
	\centering
	\caption{Effectiveness of click supervision. We exhibit the annotation cost and performance gains in both the sematic segmentation domain and the temporal action localization domain, based on a pioneering work \cite{bearman2016s} and BackTAL. It can be found that BackTAL requires less annotation cost but achieves more improvements.}
	\begin{tabular}{l|cc|cc}
		\toprule
		\toprule
		\multicolumn{1}{r}{} & \multicolumn{2}{c|}{Semantic Segmenation} & \multicolumn{2}{c}{Action Localization} \\
		\midrule
		\midrule
		\multicolumn{5}{c}{Annotation Cost} \\
		\midrule
		Corresponding SOTA & \cite{papandreou2015weakly} & 67 h  & \cite{moniruzzaman2020action} & 45 s \\
		Click Supervision & \cite{bearman2016s} & 79 h  & BackTAL & 48 s \\
		Relative Improvement & -     & 17.9\% & -     & 6.7\% \\
		\midrule
		\midrule
		\multicolumn{5}{c}{Performance Gains} \\
		\midrule
		Corresponding SOTA & \cite{papandreou2015weakly} & 39.6  & \cite{moniruzzaman2020action} & 32.3 \\
		Click Supervision & \cite{bearman2016s} & 43.6  & BackTAL & 36.3 \\
		Relative Improvement & -     & 10.1\% & -     & 12.4\% \\
		\bottomrule
		\bottomrule
	\end{tabular}%
	\label{tab-trade-off}%
\end{table}%

\textit{Further discussions.} There may be one concern about the trade-off between annotation costs and performance gains. As shown in Fig. \ref{fig-trade-off}, we compare the proposed BackTAL with recent works. It can be found that the background click supervision requires similar annotation cost (48s \textit{v.s.} 45s) with traditional weakly supervised methods, but can steadily improve the performance from 33.7 mAP (reported by Lee \etal \cite{lee2021weakly}) to 36.3 mAP. Besides, we analyze the effectiveness of the click-level supervision in both the semantic segmentation domain and the action localization domain, based on a pioneering work \cite{bearman2016s} and the proposed BackTAL. As shown in Table \ref{tab-trade-off}, compared with corresponding state-of-the-art method \cite{papandreou2015weakly}, Bearman \etal \cite{bearman2016s} require 17.9\% extra annotation cost and make 10.1\% relative improvement. In comparison, BackTAL achieves 12.4\% relative improvements while the extra annotation cost is 6.7\%\footnote{Following Bearman \etal \cite{bearman2016s}, we compare with state-of-the-art work \cite{moniruzzaman2020action} published in the previous year.}. From above analysis, we can find the effectiveness of the proposed BackTAL method, especially the good trade-off between annotation costs and performance gains.

\begin{figure*}[t]
	\graphicspath{{figure/}}
	\centering
	\includegraphics[width=1\linewidth]{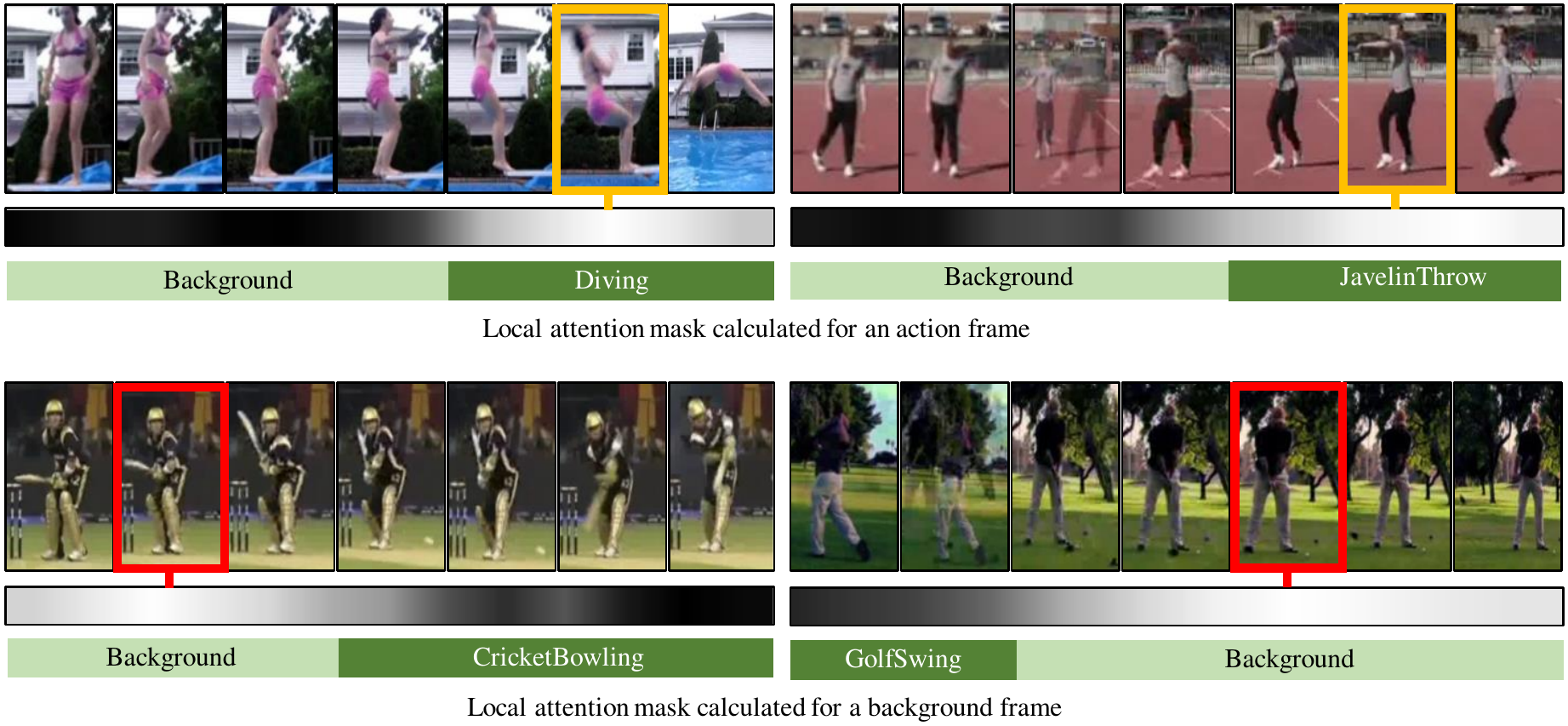}
	\caption{Visualization of the local attention mask. For each example, we show the attention mask calculated between a selected action frame (shown in orange) or a background frame (shown in red) and its corresponding neighboring frames.}
	\label{fig:similarity-exp}
\end{figure*}

\subsection{Ablation Studies}
\label{section-ablation}

\textbf{Superiority of annotating backgrounds.} We compare the proposed background-click annotation with previous action-click annotation proposed by SF-Net \cite{ma2020sf}, and report the results in Table \ref{ann-bg}. We adopt the action-click annotation released by SF-Net \cite{ma2020sf} for fair comparison. Starting from the same baseline method, introducing action-click supervision brings 0.5 mAP improvement, while the proposed background-click supervision brings 6.4 mAP improvements. When only action-click supervision or background-click supervision is available, apart from the video-level classification loss used by the baseline method, we only introduce the frame-level classification loss on the annotated frame, and do not employ any other loss functions. This demonstrates our assumption that the background-click annotation is more valuable than the action-click one, because representative action frames can be discovered by the top-$k$ aggregation process and the majority of localization errors come from the \textit{Background Error}. Moreover, starting from \textit{Baseline + Action Click}, introducing the background-click annotation can still improve the performance from 29.1 mAP to 36.8 mAP. This demonstrates the background-click annotation are quite complementary to the action-click annotation. In contrast, starting from \textit{Baseline + Background Click}, introducing the action-click annotation only improves the performance from 35.0 mAP to 36.8 mAP, which is consistent with our hypothesis that the action-click annotation are redundant with the top-$k$ aggregation process to some extent. It is worth noting that the performance gains brought by the score separation module and the affinity module are comparable to some recent works \cite{ma2020sf, liu2021weakly}.

\begin{table}[t]
	\centering
	\caption{Comparison of the background-click annotation with the action-click annotation on THUMOS14 dataset.}
	\begin{threeparttable}[t]
		\begin{tabular}{l|c}
			\toprule
			\toprule
			Setting & mAP@tIoU0.5 (\%) \\
			\midrule
			\midrule
			Baseline & 28.6 \\
			Baseline + Action Click & 29.1 \\
			Baseline + Background Click & 35.0 \\
			Baseline + Action \& Background Click & 36.8 \\
			\bottomrule
			\bottomrule
		\end{tabular}%
		\begin{tablenotes}
			\item \textit{The action-click annotation is from SF-Net \cite{ma2020sf}.}
		\end{tablenotes}
	\end{threeparttable}
	\label{ann-bg}%
\end{table}%

\begin{table}[t]
	\centering
	\caption{Ablation studies about the efficacy of the score separation module and the affinity module on THUMOS14 dataset.}
	\begin{tabular}{cccc|c}
    \toprule
    \toprule
    \multirow{2}[2]{*}{Baseline} & Background & Score & Affinity & mAP@ \\
          & Click & Separation & Module & tIoU0.5(\%) \\
    \midrule
    \midrule
    \checkmark &       &       &       & 28.6 \\
    \checkmark & \checkmark &       &       & 35.0 \\
    \checkmark & \checkmark & \checkmark &       & 35.6 \\
    \checkmark & \checkmark &       & \checkmark & 35.8 \\
    \checkmark & \checkmark & \checkmark & \checkmark & 36.3 \\
    \bottomrule
    \bottomrule
    \end{tabular}%
	\label{tab:ablation-components}%
\end{table}%

\begin{figure*}[t]
	\graphicspath{{figure/}}
	\centering
	\includegraphics[width=1\linewidth]{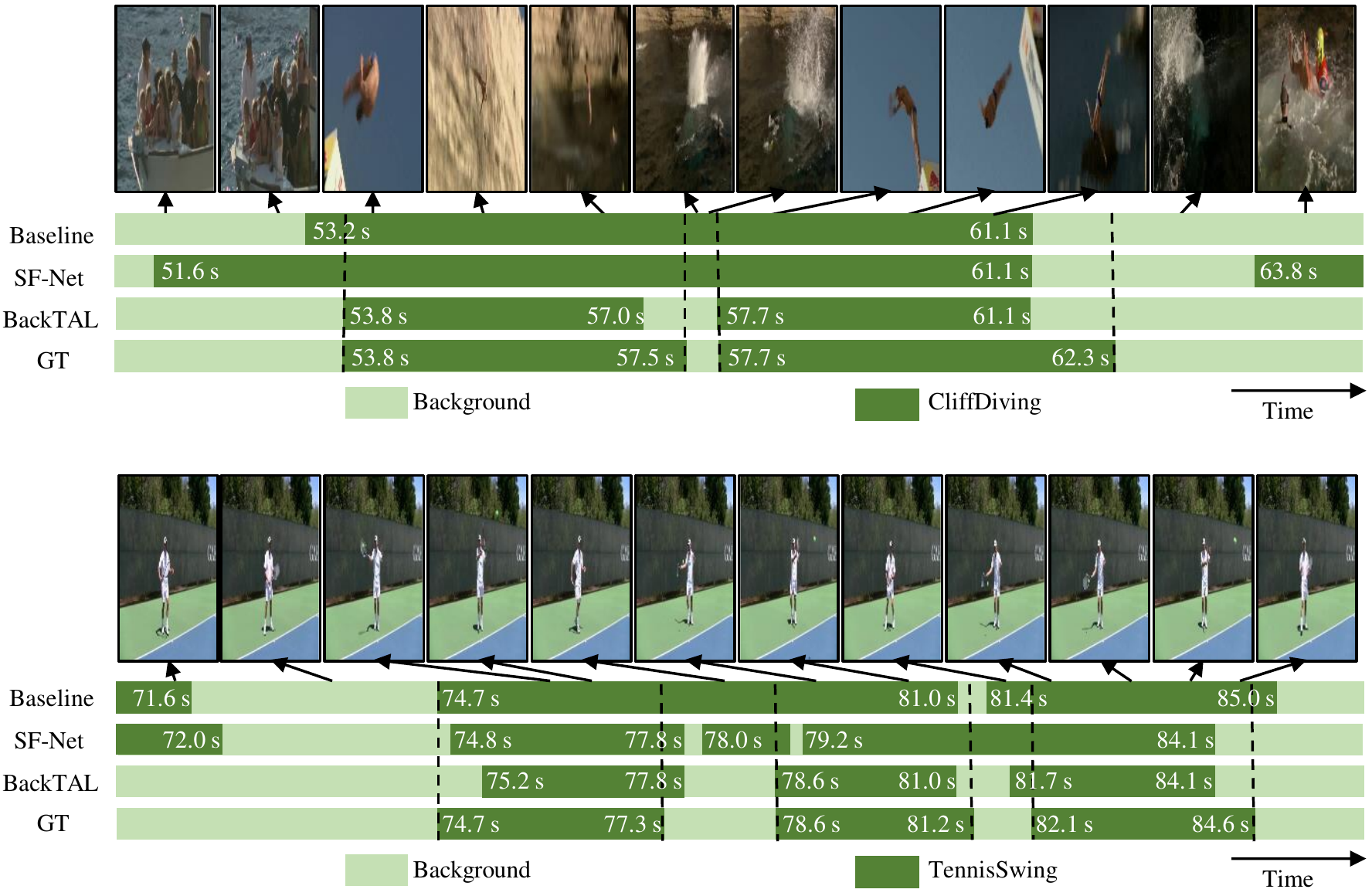}
	\caption{Qualitative comparisons between the proposed BackTAL and SF-Net \cite{ma2020sf} on THUMOS14 dataset, where the start time and end time for each action instance is depicted. For the second visualization, please view in zoom and pay attention to the tennis ball to distinguish action frames from backgrounds.}
	\label{fig:visualization}
\end{figure*}

\textbf{Effectiveness of each module.} Table \ref{tab:ablation-components} reports ablation studies about each module. Specifically, although the official implementation of BaS-Net \cite{lee2020background} gets 27.0 mAP under tIoU threshold 0.5, we achieve 28.6 mAP via simplifying the data augmentation procedure. The background-click annotation brings obvious performance improvements and achieves 35.0 mAP. Based on this, the score separation module and the affinity module bring 0.6 mAP and 0.8 mAP improvement, respectively. In the end, the complete BackTAL method achieves 36.3 mAP. There may exist a concern that the score separation module and the affinity module do not bring obvious improvements as the background-click annotation. For one thing, the core contribution of this work is to convert action-click supervision to background-click supervision, which achieves noticeable performance gains. For another, starting from a well-performed method, the score separation module and the affinity module can further make improvements and contribute 1.3 mAP gains in total, which identifies their effectiveness. In addition, we study the influence of the affinity loss $\mathcal{L}_{\rm aff}$ by removing it from the affinity module. This experiment obtains 35.1 mAP, and verifies that removing the affinity loss would make the affinity module lose efficacy. To our best knowledge, this is due to that the insufficient supervision would cause low-quality local attention masks.

\begin{table}[t]
	\centering
	\caption{Ablation studies about mining position information in different manners. Directly mining the position information is to perform supervised classification on attention weight (weight supervision) or on class activation sequence (CAS supervision). Moreover, we propose the score separation module to further mine the position information. Experiments are performed on THUMOS14 dataset.}
	\begin{tabular}{l|c}
		\toprule
		\toprule
		\multicolumn{1}{c|}{Setting} & mAP@tIou0.5 (\%) \\
		\midrule
		\midrule
		Baseline & 28.6 \\
		\midrule
		Baseline + Weight Supervision & 34.1 \\
		Baseline + CAS Supervision & 35.0 \\
		Baseline + Weight Supervision + CAS Supervision & 35.2 \\
		\midrule
		Baseline + CAS Supervision+ Score Separation & 35.6 \\
		\bottomrule
		\bottomrule
	\end{tabular}%
	\label{mine-position}%
\end{table}%

\textbf{Different ways to mine the position information.}
\label{mining-position-information}
Given the background-click annotation, a natural choice to mine the position information is performing supervised classification on the class activation sequence. Besides, as the network learns a class-agnostic attention weight to filter out backgrounds,
\begin{table}[thbp]
  \centering
  \caption{Ablation studies about the influence of the neighboring frame number, measured by mAP (\%) under IoU threshold 0.5 on THUMOS14 dataset.}
    \begin{tabular}{l|cccc}
    \toprule
    \toprule
    Neighboring frame number & 3     & 5     & 7     & 9 \\
    \midrule
    mAP@tIoU0.5 (\%) & 36.3  & 36.1  & 36.0  & 35.8 \\
    \bottomrule
    \bottomrule
    \end{tabular}%
  \label{tabAblNeighFrame}%
\end{table}%
we can apply supervision to attention weight via performing binary classification. Moreover, we can jointly mine the position information on both class activation sequence and attention weights. Experimental results are reported in Table \ref{mine-position}, under tIoU threshold 0.5. First of all, mining the position information brings adequate performance gains over the baseline method. To be specific, ``CAS Supervision" performs better than ``Weight Supervision", but simultaneously utilizing these two kinds of supervision cannot obviously exhibit further improvement. This demonstrates multiple variants of simple frame-wise classification are coessential and cannot additively improve the localization performance. In contrast, the proposed score separation module explicitly models responses of actions and backgrounds. The target to enlarge the score gap lifts the response for action frames and suppresses the response for backgrounds, which further improves the performance from 35.0 mAP to 35.6 mAP.

\textbf{Ablations about the number of neighboring frames.} In the affinity module, we keep the same value between the number of neighboring frames $h$ and the size of the temporal convolutional kernel. Alternatively, we can first calculate the weighted sum of $h$ neighboring frames, then perform the temporal convolution. As shown in Table \ref{tabAblNeighFrame}, we do not observe performance improvement when varying $h$ from 3 to 9. Because the number of neighboring frames influences the scope of context, one potential reason is that a proper context (\eg $h=3$) can enhance the feature representation, while excessive context would bring unnecessary noise.

\begin{table}[thbp]
  \centering
  \caption{Complexity comparison between our BackTAL and recent action localization methods, in terms of model parameters (M) and computational FLOPs (G).}
  \footnotesize
  \setlength{\tabcolsep}{2.0pt}
    \begin{tabular}{llllll}
    \toprule
    \toprule
    \multicolumn{1}{l|}{Research} & \multicolumn{1}{c}{3C-Net \cite{narayan20193c}} & \multicolumn{1}{c}{SF-Net \cite{ma2020sf}} & \multicolumn{1}{c}{UM \cite{lee2021weakly}} & \multicolumn{1}{c|}{{\scriptsize HAM-Net} \cite{islam2021a}} & \multicolumn{1}{c}{BackTAL} \\
    \multicolumn{1}{l|}{Publication} & \multicolumn{1}{c}{ICCV 2019} & \multicolumn{1}{c}{ECCV 2020} & \multicolumn{1}{c}{{\scriptsize AAAI 2021}} & \multicolumn{1}{c|}{AAAI 2021} & \multicolumn{1}{c}{-} \\
    \midrule
    \multicolumn{1}{l|}{Para.} & \multicolumn{1}{c}{4.41} & \multicolumn{1}{c}{16.83} & \multicolumn{1}{c}{12.63} & \multicolumn{1}{c|}{29.15} & \multicolumn{1}{c}{4.29} \\
    \multicolumn{1}{l|}{FLOPs} & \multicolumn{1}{c}{6.60} & \multicolumn{1}{c}{25.24} & \multicolumn{1}{c}{18.94} & \multicolumn{1}{c|}{43.73} & \multicolumn{1}{c}{6.51} \\
    \midrule
    \midrule
    \multicolumn{6}{l}{``Para." indicates model parameters.} \\
    \end{tabular}%
  \label{tabMedComplex}%
\end{table}%

\textbf{Computational complexity.} Table \ref{tabMedComplex} compares the computational complexity in terms of model parameters and computational FLOPs. As can be seen, our approach has lower computational complexity than recent methods SF-Net \cite{ma2020sf}, UM \cite{lee2021weakly}, and HAM-Net \cite{islam2021a}. Notably, compared to the most recent method HAM-Net \cite{islam2021a}, our BackTAL only has 14.72\% of its parameters and 14.89\% of its FLOPs.

\textbf{Dimension of embedding.}
In the affinity module, BackTAL learns an embedding for each frame with the target of distinguishing action frames from background frames. Considering different embedding dimensions lead to different representation ability of the embedding vector, we carry ablation experiments to study the influence of embedding dimension $D_{\rm emb}$. As reported in Table \ref{embedding-dim}, BackTAL achieves high performance 36.3 mAP when $D_{\rm emb}$=32. Smaller embedding dimensions may constrain the representation ability, while larger embedding dimensions are difficult to learn, which constrains the performance of BackTAL.

\begin{table}[t]
	\centering
	\caption{Exploration of different embedding dimensions for the temporal action localization performance on THUMOS14 dataset.}
	\begin{tabular}{l|ccccc}
		\toprule
		\toprule
		Embedding Dimension & 8     & 16    & 32    & 64    & 128 \\
		\midrule
		mAP@tIoU0.5 (\%) & 35.7  & 35.9  & 36.3  & 36.2  & 35.9 \\
		\bottomrule
		\bottomrule
	\end{tabular}%
	\label{embedding-dim}%
\end{table}%

\begin{table}[t]
	\centering
	\caption{Ablation studies about the influence of four hyper-parameters: the balance coefficients in the complete loss function $\lambda$ and $\beta$, thresholds $\tau_{\rm same}$ and $\tau_{\rm diff}$ to calculate the embedding loss on THUMOS14 dataset.}
	\begin{tabular}{c|ccc}
		\toprule
		\toprule
		$\lambda$ & 0.8   & \cellcolor[rgb]{ .906,  .902,  .902}1.0 & 1.2 \\
		mAP@tIoU0.5 (\%) & 35.8  & \cellcolor[rgb]{ .906,  .902,  .902}36.3 & 36.2 \\
		\midrule
		$\beta$ & 0.6   & \cellcolor[rgb]{ .906,  .902,  .902}0.8 & 1.0 \\
		mAP@tIoU0.5 (\%) & 35.9  & \cellcolor[rgb]{ .906,  .902,  .902}36.3 & 36.2 \\
		\midrule
		$\tau_{\rm same}$ & 0.3   & \cellcolor[rgb]{ .906,  .902,  .902}0.5 & 0.7 \\
		mAP@tIoU0.5 (\%) & 36.1  & \cellcolor[rgb]{ .906,  .902,  .902}36.3 & 35.6 \\
		\midrule
		$\tau_{\rm diff}$ & 0.0   & \cellcolor[rgb]{ .906,  .902,  .902}0.1 & 0.2 \\
		mAP@tIoU0.5 (\%) & 35.6  & \cellcolor[rgb]{ .906,  .902,  .902}36.3 & 36.0 \\
		\bottomrule
		\bottomrule
	\end{tabular}%
	\label{tabCoefExp}%
\end{table}%

\begin{table}[thbp]
	\centering
	\caption{Ablation studies about the efficacy of the score separation module and the affinity module on the THUMOS14 dataset, based on the action-click annotation and mined background frames.}
	\setlength{\tabcolsep}{4.0pt}
	\begin{tabular}{ccccc|c}
    \toprule
    \toprule
    \multirow{2}[2]{*}{Baseline} & Action & Mined & Score & Affinity & mAP@ \\
          & Click & Bg. Frames & Separation & Module & tIoU0.5(\%) \\
    \midrule
    \midrule
    \checkmark &       &       &       &       & 28.6 \\
    \checkmark & \checkmark &       &       &       & 29.1 \\
    \checkmark & \checkmark & \checkmark &       &       & 30.2 \\
    \checkmark & \checkmark & \checkmark & \checkmark &       & 31.6 \\
    \checkmark & \checkmark & \checkmark &       & \checkmark & 31.8 \\
    \checkmark & \checkmark & \checkmark & \checkmark & \checkmark & 32.4 \\
    \bottomrule
    \bottomrule
    \end{tabular}%
	\label{tabAblActionClickModule}%
\end{table}%

\textbf{Influence of hyper-parameters.} In the proposed BackTAL, the balance coefficients $\lambda$ and $\beta$, thresholds $\tau_{\rm same}$ and $\tau_{\rm diff}$ are empirically determined. We carry ablation experiments to study the influence of these hyper-parameters. Specifically, we change one hyper-parameter when fixing others, and verify temporal action localization performance on THUMOS14 dataset. As shown in Table \ref{tabCoefExp}, when hyper-parameters change in a reasonable range, we can observe a certain performance variation. For example, decreasing the coefficients $\lambda$ in the loss function would drop the 0.5 mAP performance. Increasing $\tau_{\rm same}$ would make the algorithm to select similar action (or background) frames more strict. Consequently, BackTAL would select less vectors to learn the embedding space, which damages the performance. Similar tendency can be found for decreasing the threshold $\tau_{\rm diff}$. In contrast, decreasing $\tau_{\rm same}$ or increasing $\tau_{\rm diff}$ would guide BackTAL to select more vectors to learn the embedding space. The redundant embedding vectors may bring noises to the learning process and constrain the performance.

\textbf{Performance based on the action-click annotation.} Moreover, we use the action-click annotation of SF-Net \cite{ma2020sf} and adopt SF-Net's strategy to mine background frames. This experimental result obtains 32.4 mAP, as shown in Table \ref{tabAblActionClickModule}. On the one hand, owing to the proposed score separation module and affinity module, our BackTAL (32.4 mAP) exceeds SF-Net (30.5 mAP) when using the same action-click supervision. On the other hand, the performance gap between the action-click based method (32.4 mAP) and background-click based BackTAL (36.3 mAP) demonstrates the effectiveness of the background-click supervision.

\subsection{Qualitative Analysis}
\label{sec-qual-exps}
This section analyzes the proposed BackTAL method in qualitative manner. First of all, Fig. \ref{fig:similarity-exp} visualizes the local attention mask employed in the affinity module. It can be found that, given an action frame, the local attention mask can highlight neighboring action frames and suppress background frames, and vice versa. Based on this, the local attention mask serves as the frame-specific attention weight and guides the calculation of temporal convolution. In the end, high-quality local-attention mask assists in generating discriminative class activation sequence.

Besides, Fig. \ref{fig:visualization} compares the proposed BackTAL with the baseline method and the strong competitor SF-Net \cite{ma2020sf}. Both the baseline method and SF-Net take some risks to improperly regard confusing background frames as actions. For example, the people surfaced after diving may be regarded as a part of \textit{CliffDiving} action. The hand moving, but not the complete swing action, can be regarded as a \textit{TennisSwing} action. Because the insufficient ability to suppress confusing background frames, the algorithm may regard multiple adjacent action instances as a long action instance, or localize imprecise action boundaries. In contrast, the proposed BackTAL can consistently suppress confusing background frames and precisely separate adjacent action instances. In experiments, we also notice that BackTAL breaks some long action instances into several separated instances. These failure cases occur when there is extreme variations within the action instance. For example, the viewpoint change can cause extreme variation about object size. These failure cases remind that the weakly supervised temporal action localization should be further developed.

\section{Conclusion}
\label{sec-conclusion}
We develop the action-click supervision into the background-click supervision, and propose BackTAL for weakly supervised temporal action localization. We cast the learning process as mining both the position information and the feature information, and propose the score separation module and the affinity module, to mitigate the action-context confusion challenge. In experiments, BackTAL builds new high performance on two traditional benchmarks, \ie THUMOS14 and ActivityNet v1.2, and reports a promising performance on a recent large-scale benchmark, HACS. Moreover, we verify the efficacies of explicitly separating action scores and background scores, as well as dynamically attending to informative neighbors. In the future, we plan to introduce the spirit of background-click supervision to similar weakly supervised learning domain, \eg weakly supervised object localization \cite{guo2021strengthen} and detection \cite{zhang2020weakly}, pointly-supervised semantic segmentation \cite{bearman2016s}. Besides, it is promising to study the inherent correlations between position information and feature information to further develop the background-click supervision.

\bibliographystyle{IEEEtran}
\bibliography{egbib}

\ifCLASSOPTIONcaptionsoff
  \newpage
\fi

\begin{IEEEbiography}[{\includegraphics[width=1in,height=1.25in,clip,keepaspectratio]{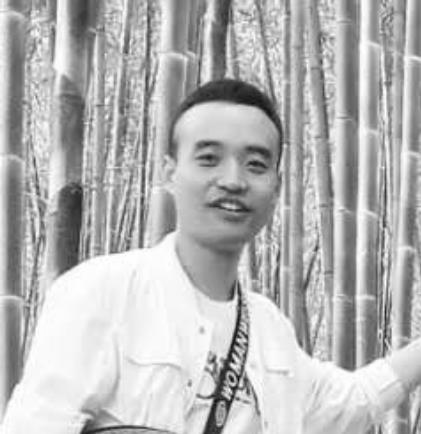}}]{Le Yang} received his B.E. degree from Northwestern Polytechnical University, Xi'an, China, in 2016. He is currently a Ph.D. candidate in the School of Automation at Northwestern Polytechnical University. His research interests include temporal action localization, video object segmentation and weakly supervised learning.
\end{IEEEbiography}

\begin{IEEEbiography}[{\includegraphics[width=1in,height=1.25in,clip,keepaspectratio]{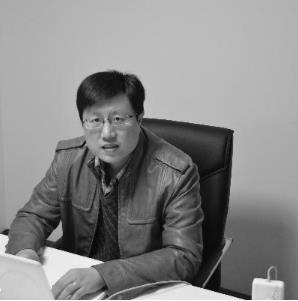}}]{Junwei Han}
	is currently a Professor in the School of Automation, Northwestern Polytechnical University. His research interests include computer vision, pattern recognition, remote sensing image analysis, and brain imaging analysis. He has published more than 70 papers in top journals such as IEEE TPAMI, TNNLS, IJCV, and more than 30 papers in top conferences such as CVPR, ICCV, MICCAI, and IJCAI. He is an Associate Editor for several journals such as IEEE TNNLS and IEEE TMM.
\end{IEEEbiography}

\begin{IEEEbiography}[{\includegraphics[width=1in,height=1.25in,clip,keepaspectratio]{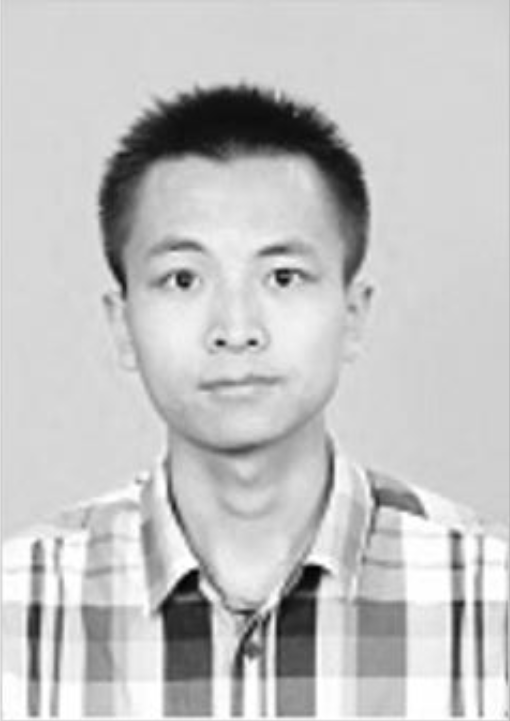}}]{Tao Zhao} received his M.S. degree from Northwestern Polytechnical University, Xi'an, China, in 2018. He is currently a Ph.D. candidate in the School of Automation at Northwestern Polytechnical University. His research interests include video temporal action localization and weakly supervised learning.
\end{IEEEbiography}

\begin{IEEEbiography}[{\includegraphics[width=1in,height=1.25in,clip,keepaspectratio]{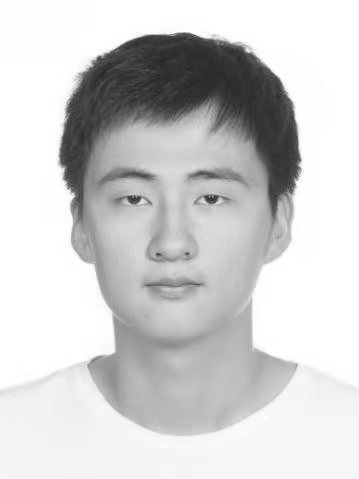}}]{Tianwei Lin}  received his master degree at Shanghai Jiao Tong University in 2019, advised by Prof. Xu Zhao. Tianwei Lin received his B.Eng from School of Mechanical Engineering at Shanghai Jiao Tong University in 2016. His research interests include: Computer Vision, Deep Learning, Action Recognition, Temporal Action Detection, GAN.
\end{IEEEbiography}

\begin{IEEEbiography}[{\includegraphics[width=1in,height=1.25in,clip,keepaspectratio]{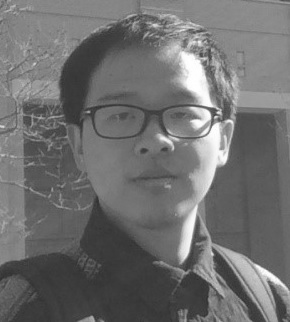}}]{Dingwen Zhang} received his Ph.D. degree from the Northwestern Polytechnical University, Xi'an, China, in 2018. He is currently a professor in the Brain Lab, Northwestern Polytechnical University. From 2015 to 2017, he was a visiting scholar at the Robotic Institute, Carnegie Mellon University. His research interests include computer vision and multimedia processing, especially on saliency detection, video object segmentation, and weakly supervised learning.
\end{IEEEbiography}

\begin{IEEEbiography}[{\includegraphics[width=1in,height=1.25in,clip,keepaspectratio]{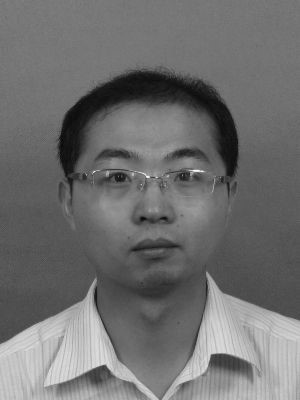}}]{Jianxin Chen} is Professor of artificial intelligence and machine learning at Beijing University of Chinese Medicine(BUCM),where he leads the Cognitive Group within the Center for Vision, Speech and Signal Processing about Traditional Chinese Medicine(TCM). His research centers on the use of artificial intelligence to elaborate TCM. He is an associate editor for Pharmacological Research.
\end{IEEEbiography}

\end{document}